\documentclass[12pt]{article}

\usepackage[margin=1in]{geometry}
\usepackage{setspace}
\usepackage{graphicx}
\usepackage{amsmath,amssymb}
\usepackage{booktabs}
\usepackage{caption}
\usepackage{subcaption}
\usepackage{hyperref}
\usepackage{algorithm}
\usepackage{algpseudocode}
\usepackage{tabularx} 
\usepackage[table]{xcolor} 

\newcommand{\keywords}[1]{%
  \par\noindent\textbf{Keywords:} #1
}

\begin{document}

\title{Robust and Clinically Reliable EEG Biomarkers: A Cross-Population Framework for Generalizable Parkinson’s Disease Detection}
\author{ \small
Nicholas R. Rasmussen$^{1}$,
Longwei Wang$^{1}$,
Rodrigue Rizk$^{1}$,
Md Rezwanul Akter Pallab$^{2}$,\\ \small
Samuel Stuwart$^{3}$,
Martina Mancini$^{4}$,
Arun Singh$^{2}$,
KC Santosh$^{1}$ \\ \small
$^{1}$USD AI Research, Department of Computer Science,\\ \small University of South Dakota, Vermillion, SD 57069, USA \\ 
\small
$^{2}$Biomedical and Translational Sciences, Sanford School of Medicine,\\ \small University of South Dakota, Vermillion, SD 57069, USA \\ \small
$^{3}$Department of Sport, Exercise and Rehabilitation, Northumbria University,\\
\small Newcastle upon Tyne, NE1 8ST, United Kingdom \\
\small
$^{4}$Department of Neurology, Oregon Health \& Science University,\\ \small Portland, OR 97239, USA \\ \small
\{nicholas.rasmussen, rezwan.pallab\}@coyotes.usd.edu,
\{rodrigue.rizk, longwei.wang, \\ \small arun.singh, kc.santosh\}@usd.edu,
mancinim@ohsu.edu, samuel2.stuart@northumbria.ac.uk
}

\maketitle

\section*{Statements and Declarations}

\subsection*{Funding}
This work was supported by the National Science Foundation under Grant No.~\href{https://www.nsf.gov/awardsearch/showAward?AWD_ID=2346643}{\#2346643}, the U.S. Department of Defense under Award No.~\href{https://dtic.dimensions.ai/details/grant/grant.14525543}{\#FA9550-23-1-0495}, and the U.S. Department of Education under Grant No.~P116Z240151. Computations supporting this project were performed on High Performance Computing systems at the University of South Dakota, funded by NSF Award OAC‑1626516. This work was supported in part by the Parkinson’s Foundation Fellowship for Basic Scientists awarded to Dr.~Samuel Stuart (PF‑FBS‑1898‑18‑01).

\subsection*{Author Contributions}

\textbf{Manuscript drafting, experimental design, implementation, and evaluation:} Nicholas R.~Rasmussen.

\textbf{Discussion formulation and theoretical framing:} Longwei Wang and Nicholas R.~Rasmussen.

\textbf{Data collection and cohort assembly:} Md.~Rezwanul Akter Pallab and Arun Singh (USD cohort); Samuel Stuart and Martina Mancini (OHSU cohort).

\textbf{Manuscript review and editorial feedback:} Longwei Wang, Rodrigue Rizk, Samuel Stuart, Martina Mancini, Arun Singh, and K.~C.~Santosh.

\textbf{Methodological conceptualization:} Arun Singh and K.~C.~Santosh.

\textbf{Evaluation conceptualization:} Rodrigue Rizk and K.~C.~Santosh.

\subsection*{Employment and Affiliations}
During preparation of this manuscript, Nicholas R.~Rasmussen was employed by the University of Washington and volunteered with The COVID Detection Foundation (DBA Virufy).

\subsection*{Ethics Approval and Data Availability}

All data analyzed in this study were previously collected and reported in prior publications. Ethical approval and informed consent procedures for each dataset were obtained as described in the corresponding original studies.

The USD dataset was approved by the local institutional ethics committee, and all participants provided written informed consent prior to data collection.

All other datasets used in this work were obtained from their institutions or online repositories and were analyzed in accordance with the terms and conditions specified by their original data providers. No additional ethical approval was required for secondary analysis of these de-identified, publicly released datasets.

\clearpage

\begin{abstract}

Developing robust and clinically reliable EEG biomarkers requires evaluation frameworks that explicitly address cross-population generalization in multi-site settings such as Parkinson’s disease (PD) detection. Models trained under i.i.d. assumptions often capture population-specific artifacts rather than disease-relevant neural structure, leading to poor generalization across clinical cohorts. EEG further amplifies this challenge due to low signal-to-noise ratio and heterogeneous acquisition conditions. We propose a population-aware evaluation framework to assess the robustness and clinical reliability of EEG biomarkers under distribution shift. Using an $n$-gram expansion strategy, we enumerate all cross-population train--test configurations across five independent cohorts, resulting in 75 directional evaluations. A nested cross-validation design with integrated channel selection ensures prospective biomarker identification without population leakage. Results show that cross-population transfer is asymmetric and that both accuracy and biomarker stability improve with increasing training-population diversity, achieving up to $94.1\%$ accuracy on held-out cohorts. A theoretical analysis based on mixture-risk optimization and hypothesis-space contraction explains these trends, showing that multi-population training promotes population-robust representations. This work establishes a principled framework for learning robust, generalizable, and clinically reliable EEG biomarkers for multi-site biomedical applications.

\end{abstract}

\keywords{Biomarker extraction; Distribution shift; Cross‑population generalization; EEG signal classification; Channel selection; Multi‑site evaluation}

\section{Introduction}
\label{intro}

\begin{figure}[h]
    \centering
    \includegraphics[width=.495\linewidth]{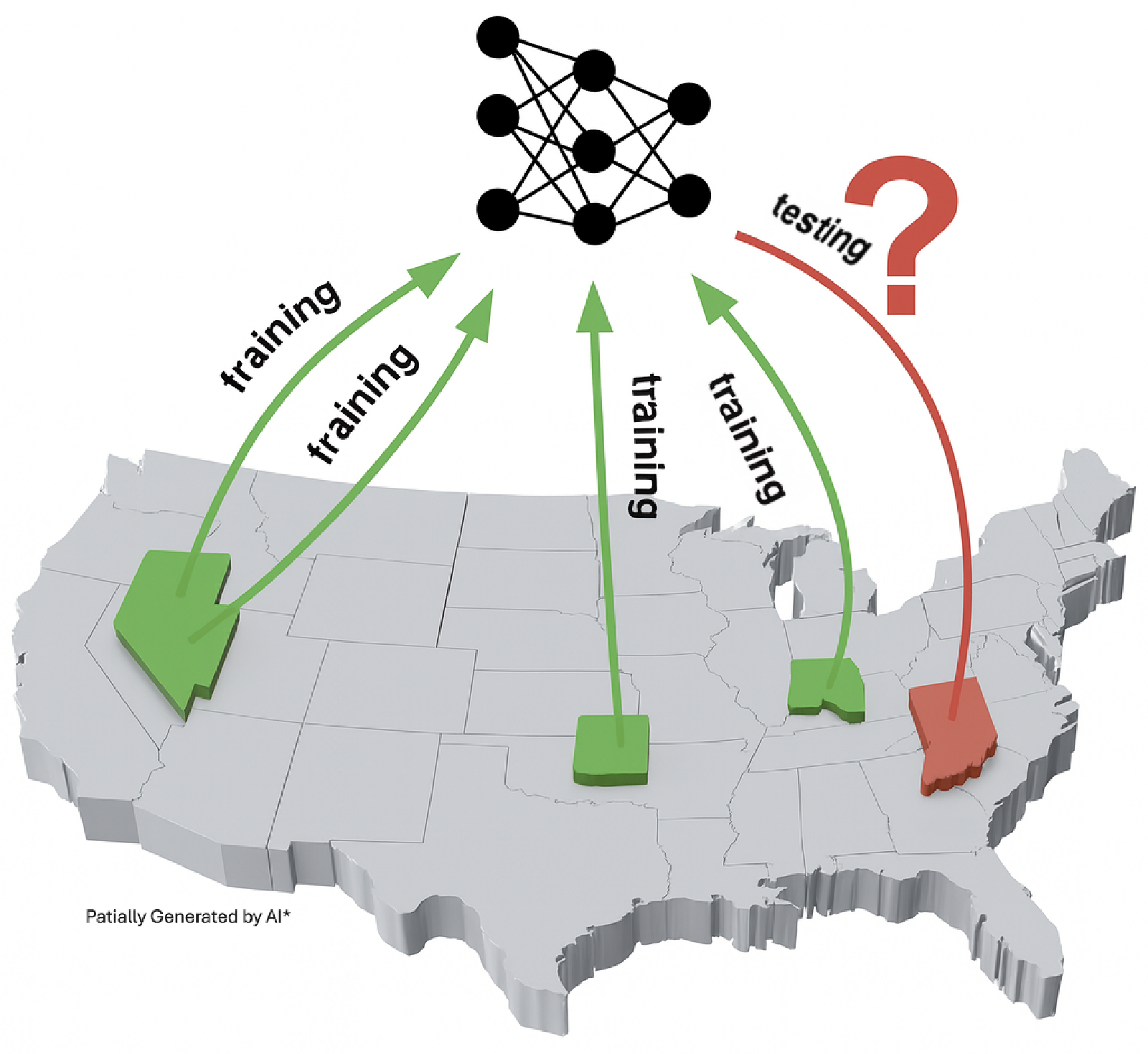}
    \caption{ Conceptual illustration of cross‑population generalization in ML‑based PD detection. Training on multiple regional cohorts (green) does not guarantee reliable performance on an unseen population (red), highlighting the need for rigorous cross‑population evaluation in clinical ML.}
    \label{fig:motivation}
\end{figure}

Machine learning (ML) approaches to biomedical signal analysis increasingly aim to extract reproducible biomarkers, yet the independent‑and‑identically‑distributed (i.i.d.) assumption underlying standard empirical risk minimization rarely holds when data originate from heterogeneous populations, institutions, and acquisition pipelines \cite{shalev2014understanding, vapnik1998statistical, quinonero2009dataset}. Violations of this assumption introduce distribution shift that can distort biomarkers and performance estimates masking failures that remain invisible under within‑dataset validation \cite{bradshaw2023crossvalidation, geirhos2020shortcut}. In clinical contexts, instability in biomarker extraction carries consequences beyond reduced accuracy, including misdiagnosis, delayed treatment, and inappropriate clinical decisions \cite{zech2018variable, compton2023when, cross2024bias, roberts2021common}. As illustrated in Fig.~\ref{fig:motivation}, strong retrospective performance on a single dataset does not guarantee that the extracted signal reflects disease‑relevant physiology rather than population‑specific artifacts \cite{avola2025benchmarking}. These observations motivate a computational research agenda centered on evaluation frameworks that explicitly characterize biomarker robustness under cross‑population transfer while not relying on i.i.d. proxies.

Electroencephalography (EEG) presents an especially diffcult instance of this problem. EEG cohorts collected across sites differ in hardware manufacturers, amplifier characteristics, sampling rates, electrode counts, electrode layouts, acquisition protocols, and medication states \cite{anjum2024resting, may2023portlandEEG, aljalal2022parkinson}. Even nominally identical systems can produce systematic distributional differences due to electrode placement variability, impedance, and technician practice \cite{zech2018variable}. Combined with the low signal-to-noise ratio intrinsic to scalp EEG, these acquisition-level variations create a setting in which models can readily learn cohort-specific structure, site-specific artifacts, or demographic confounds, rather than task-relevant neurophysiological signatures \cite{roy2019deep, geirhos2020shortcut}. While channel harmonization and spatial normalization can attenuate some sources of variability, however residual discrepancies persist as irreducible obstacles to cross-site transfer that confound and obscure the relevant information of the diseases we aim to capture.

Recent studies applying ML to EEG-based Parkinson's disease (PD) detection have reported strong within-dataset classification performance \cite{anjum2024resting, may2023portlandEEG, rockhill2020ucsdEEG, singh2018parkinsons, singh2020frontal}. However, these results are computationally fragile for several reasons. First, many evaluations rely on non-nested cross-validation protocols in which model selection and hyperparameter tuning are not strictly separated from test-set evaluation, permitting information leakage and optimistic bias \cite{brookshire2024data, roberts2017cross, schratz2019hyperparameter}. Second, single-site evaluation conflates model capacity with distributional coverage: high accuracy on one cohort does not distinguish disease-relevant features from site-specific shortcuts \cite{geirhos2020shortcut, shafiezadeh2023methodological}. Third, aggregate performance metrics obscure population-specific failure modes; complementary analyses such as directional transfer matrices and precision--recall decompositions are rarely reported \cite{roberts2021common}. These limitations of the literature mean that existing performance claims rest on evaluation protocols that overestimate deployability under heterogeneous, real-world acquisition conditions.

Overestimation is not merely a reporting gap, as standard domain-generalization theory demonstrates that minimizing empirical risk on a source mixture does not guarantee low risk on a held-out target population unless the target lies near the convex hull of the training mixture \cite{ben2010theory, mansour2008domain}. Cross-population transfer is therefore inherently directional and non-reciprocal: performance from cohort $A$ to cohort $B$ is structurally independent of performance from $B$ to $A$ \cite{torralba2011unbiased, gulrajani2020search}. Evaluation frameworks that pool populations or report only symmetric, averaged accuracy collapse this directional structure and suppress the very information needed to assess whether a model has learned generalizable representations or site-specific artifacts. Addressing this gap requires computational infrastructure that preserves population identity, enumerates directional transfer, and provides formal grounding for when and why generalization succeeds or fails.

Our work introduces a population‑aware computational framework for evaluating biomarker robustness under cross‑dataset distribution shift, applied to five independently acquired EEG cohorts \( (N = 285)\). Our contributions are:

\begin{itemize}
    \item A demonstration of directional, non‑reciprocal biomarker transfer across populations.

    \item An n‑gram evaluation framework enumerating all train--test configurations preserving block structure.

    \item A formal mixture‑risk analysis via hypothesis‑space contraction revealing structural multi‑population geometry.

    \item A channel‑knowledge extraction procedure for robust spatial biomarkers characterized as geometric projection.
\end{itemize}

The remainder of this paper is organized as follows. Section~\ref{ReWo} reviews prior work on generalization, multi-site evaluation methodology, and EEG-based PD analysis. Section~\ref{MnM} describes the datasets, preprocessing pipeline, embedding framework, and experimental protocol, including the n-gram population expansion strategy and nested cross-validation design. Section~\ref{ExpRe} presents classification results under the cross-population evaluation framework. Section~\ref{S:Discussion} contextualizes these findings for a formal theoretical grounding through mixture-risk optimization, hypothesis-space contraction, and channel selection as geometric projection. Section~\ref{Lim} discusses limitations, and Section~\ref{Conc} concludes with a summary of contributions and implications for multi-site biomedical signal classification.

\section{Background}
\label{ReWo}

The computational challenges of generalization under distribution shift, outlined in the preceding section, have direct implications for how EEG-based biomarker discovery and classification systems are designed, validated, and reported. This section operationalizes those challenges by reviewing the methodological landscape of cross-population evaluation and surveying prior work in EEG-based PD biomarker detection, with attention to where existing studies satisfy, or fall short of, the evaluation criteria required for credible, population-robust biomarker extraction.

Six methodological criteria are central to credible cross-population evaluation and biomarker robustness assessment, each of which has been incompletely addressed in prior EEG-based PD work:

\begin{itemize}
    \item \textbf{Subject-level separation and leakage prevention.} Training and test partitions must be separated at the subject level to prevent leakage from repeated measurements or temporally adjacent windows, which can inflate performance estimates \cite{brookshire2024data}. Many EEG-PD studies do not report whether subject-level separation was enforced, and some explicitly segment continuous recordings into overlapping windows prior to random splitting.

    \item \textbf{Cross-dataset and held-out site evaluation.} Evaluation on independently acquired cohorts or held-out institutions includes a more realistic assessment of deployment performance than within-dataset cross-validation alone \cite{roberts2021common, shafiezadeh2023methodological}. The majority of EEG-PD studies evaluate on a single dataset or, when multiple datasets are used, pool them without preserving population identity.

    \item \textbf{Nested cross-validation and hyperparameter selection.} Proper nesting of model selection and hyperparameter tuning prevents information leakage from test folds and reduces optimistic bias \cite{roberts2017cross, schratz2019hyperparameter}. Non-nested protocols remain common in the EEG-PD literature, complicating interpretation of reported accuracy.

    \item \textbf{Metrics beyond accuracy.} Aggregate accuracy can obscure clinically relevant failure modes; complementary metrics such as precision--recall curves, calibration, and class-specific sensitivity and specificity provide a more complete characterization of model behavior \cite{saito2015precision, steyerberg2010assessing}. Moreover, most EEG-PD methodology fail to incorporate error bars such a standard deviation or other stability metrics.

    \item \textbf{Directional and population-aware reporting.} Preserving directionality in cross-population experiments (e.g., train on A, test on B, and the reverse) reveals asymmetric transfer effects that aggregated evaluations may obscure \cite{zech2018variable, wang2018deep, roberts2021common}. To our knowledge, no prior EEG-PD study reports full directional transfer matrices.

    \item \textbf{Channel harmonization and its limits.} Channel harmonization and standardized montages can reduce some variability, but residual discrepancies often persist \cite{klem1999ten, acharya2016acns}. Studies that require identical channel sets across sites implicitly restrict the populations that can be included, limiting the scope of generalization claims.
\end{itemize}

In biomedical signal modeling more broadly, distributional variability reflects both biological heterogeneity and acquisition context, introducing systematic, site-specific structure into the data \cite{zech2018variable, schirrmeister2017deep, samek2017explainable}. Models trained on a single cohort frequently fail to transfer across clinical environments, even when within-dataset performance is strong \cite{compton2023when, bradshaw2023crossvalidation}. Studies in medical imaging and physiological signal analysis have documented substantial performance degradation under cross-institutional evaluation, motivating multi-site validation protocols and explicit consideration of acquisition variability as a first-class experimental factor \cite{zech2018variable, compton2023when}.

A growing literature has applied ML to EEG-based PD biomarker detection, with many studies reporting strong within-dataset classification performance \cite{anjum2020linear, shah2020dynamical, yuvaraj2018novel, lee2019convolutional}. Some efforts have incorporated cross-site or multi-cohort evaluation \cite{wu2024multi, zhang2025gepd, anjum2020linear}, but these studies are often constrained by limited dataset availability, channel harmonization requirements, or reliance on single train--test split. When evaluated against the criteria above, reported performance gains may reflect evaluation choices, such as non-independent splits or insufficiently nested tuning, that permit leakage or optimistic bias, not genuine cross-population robustness \cite{brookshire2024data, racine2000consistent}. Additional limitations include small cohort sizes, inconsistent validation protocols, and overreliance on single summary metrics \cite{valavi2018blockcv, saito2015precision, steyerberg2010assessing}. Collectively, prior work demonstrates both the promise of EEG for PD biomarker detection and the fragility of reported performance under realistic, heterogeneous conditions \cite{gemein2020machine, roy2019deep}. Evaluating cross-population transfer in this setting requires computational infrastructure that preserves population identity, enumerates directional transfer exhaustively, nests all model selection decisions within the evaluation protocol, and provides formal grounding for when and why generalization succeeds or fails. The following section describes a framework that fulfills the requires that we set forth in the section above, thus we apply it to five independently acquired EEG cohorts \cite{anjum2024resting, may2023portlandEEG, rockhill2020ucsdEEG, singh2018parkinsons, singh2020frontal}.

\section{Computational Framework}
\label{MnM}

As stated below, our framework preserves population identity, enforces strictly prospective model selection via nested cross‑validation, and enumerates all directional cross‑population transfer paths through n‑gram expansion. Recent methodological work has emphasized the role of channel selection and population‑aware evaluation when modeling heterogeneous EEG data \cite{rasmussen2025channel}. In this context, the experimental framework described here is designed to preserve population structure, patient‑level independence, and acquisition‑specific variability throughout model training and evaluation. We first introduce the publicly available and private EEG datasets used in this study (Section \ref{dSet}), followed by a topology‑consistent channel‑alignment strategy for harmonizing heterogeneous montages (Section \ref{M:Harmonization}) and the time–frequency windowing scheme applied uniformly across datasets (Section \ref{M:TF}). We then describe the nested, patient‑level cross‑validation protocol (Section \ref{M:CV}), the convolutional neural network architecture (Section \ref{M:Model}), and the training procedure (Section \ref{M:Training}). Finally, we detail the population performance‑scaling analyses (Section \ref{M:ScalingMetrics}) and population channel‑selection maps that characterize spatial evidence across heterogeneous EEG cohorts (Section \ref{M:ChannelMaps}).

\subsection{Datasets}
\label{dSet}

Our study integrates five independently acquired EEG datasets to enable population‑aware evaluation of cross‑dataset generalization in Parkinson’s disease (PD). Demographic and clinical characteristics of all cohorts are summarized in Table~\ref{patientMetadata}. To ensure comprehensive coverage of publicly available EEG resources suitable for PD classification at the time of this study, we incorporated all datasets meeting the following criteria: resting‑state acquisition, availability of diagnostic labels, and sufficient metadata to support cross‑dataset evaluation. This resulted in the inclusion of three independently collected public datasets spanning multiple institutions and acquisition protocols. In addition to these public resources, we incorporated two privately held datasets acquired at distinct institutions, including an expanded University of Iowa cohort, an Oregon Health \& Science University (OHSU) cohort, and a University of South Dakota (USD) cohort. Our combined datasets capture substantial heterogeneity in acquisition hardware, electrode configurations, recording protocols, and cohort composition, providing a realistic testbed for evaluating cross‑population generalization. For datasets providing medication state annotations, only recordings acquired in the clinically defined \textit{on‑medication} condition were included in the primary analyses to reduce confounding effects related to dopaminergic state. All datasets were treated as independent populations throughout training and evaluation, with population identity preserved to support directional cross‑dataset analysis.

\subsubsection{University of New Mexico (UNM; P1) and University of Iowa (Iowa; P2)}

The UNM dataset (P1)~\cite{anjum2020linear} comprises EEG recordings from 27 patient with PD and 27 demographically matched healthy control participants, collected in Albuquerque, New Mexico. Recordings consist of multi‑session 64‑channel EEG data, with each subject’s session containing sequential eyes‑closed and eyes‑open resting‑state segments acquired within the same recording. For analysis, these segments were retained in their entirety and treated as a combined eyes‑open + eyes‑closed condition. EEG was recorded using sintered Ag/AgCl electrodes with a bandpass of 0.1–100 Hz and sampled at 500 Hz on a Brain Vision acquisition system. Online referencing was performed at CPz, resulting in the absence of the CPz channel. Controls were matched to patients with PD on age and sex and did not differ in education level or estimated intelligence.

The Iowa dataset (P2)~\cite{anjum2020linear} comprises EEG recordings from 14 patient with PD and 14 demographically matched control participants, collected at the University of Iowa in Iowa City, Iowa. Recordings consist of 64‑channel resting‑state EEG acquired under eyes‑open conditions, with participants instructed to maintain visual fixation and minimize movement. Data were collected using the same Brain Vision acquisition system and sintered Ag/AgCl electrodes as the UNM cohort, sampled at 500 Hz with identical bandpass characteristics. However, online referencing was performed at Pz, resulting in the absence of the Pz channel in the released recordings.

\subsubsection{Expanded Iowa}

In addition to the publicly released Iowa cohort, we include a privately held expanded version of the University of Iowa dataset, which together with the public data, comprise EEG recordings from 82 patients with PD  and 41 neurologically healthy control participants. The expanded dataset was collected under the same experimental protocol and hardware configuration as the original public release and incorporates additional resting‑state EEG recordings acquired using the same EEG system and electrode montage, preserving consistency in acquisition conditions. The expanded cohort increases subject coverage giving additional clinical annotations, including labels supporting freezing‑of‑gait (FoG$^{+/-}$) classification. Participant metadata include age, sex, diagnostic status, and medication state at the time of acquisition. All data were de‑identified prior to analysis and approved for secondary use under the same ethical framework as the original public release.

\subsubsection{University of California, San Diego (UCSD; P3)}

The UCSD dataset (P3)~\cite{swann2015elevatedsynchrony} comprises EEG recordings from 15 individuals with PD and 16 age‑ and sex‑matched neurologically healthy control participants, collected at the University of California, San Diego. Control participants were matched to the PD group on demographic variables and cognitive status. Recordings consist of clinically acquired 32‑channel EEG data obtained under standardized resting‑state conditions. Participants with PD were under the care of a movement disorders neurologist to limit patient susceptibility to falls and exhibited mild to moderate disease severity, corresponding to Hoehn and Yahr stages 2–3, with an average disease duration of approximately four to five years.

EEG recordings were obtained while participants were seated comfortably and instructed to remain relaxed with eyes open while fixating on a visual target. Data were collected for approximately three minutes using a 32‑channel BioSemi ActiveTwo system sampled at 512 Hz. Additional channels were used to monitor ocular and muscular activity to support control analyses reported in the original study. Participants with PD were recorded in both medicated and unmedicated states on separate days, with session order counterbalanced across subjects, while control participants completed a single recording session using the same protocol. All participants provided written informed consent in accordance with institutional review board approval and the Declaration of Helsinki.

\subsubsection{OHSU Dataset (P4)}

The OHSU dataset (P4)~\cite{stuart2021brain} comprises resting‑state EEG recordings from 43 patient with PD and 21 neurologically healthy control participants, collected at a single clinical site in Portland, Oregon. Recordings were acquired using a \textit{TMSi Mobita 32‑channel mobile system} arranged according to a \textit{reduced 10–20 system} configuration. Data were obtained under \textit{eyes‑open} resting conditions, with participants instructed to remain still and minimize movement, and were sampled at \textit{2000 Hz} with hardware filtering and impedance thresholds applied according to site‑specific clinical protocols. Participant metadata include age range, sex distribution, and diagnostic status determined using established clinical criteria and rating scales. Medication state at the time of recording was documented as \textit{on}, approximately one hour post‑dosage. All recordings were de‑identified prior to analysis, and data collection and secondary use were approved by the appropriate institutional review board. At the time of this study, the dataset remains privately held, with plans for controlled public release pending completion of additional documentation and governance review.

\subsubsection{University of South Dakota Dataset (USD; P5)}

The USD dataset (P5) comprises resting‑state EEG recordings from 13 patient with PD and 4 neurologically healthy control participants, collected under standardized laboratory conditions at the USD. Data were acquired using a \textit{96‑channel BrainVision system} sampled at \textit{500 Hz}. Although acquisition employed a high‑density electrode layout, analysis was restricted to channels corresponding to the standard 10–20 system, with inclusion of the inferior occipital electrode Iz when available. This restriction was applied to ensure consistent anatomical correspondence with other datasets and to avoid reliance on higher‑density extensions (e.g., 10–10 or 10–5 layouts) that were not uniformly present across cohorts. Recordings were obtained under \textit{eyes‑open} resting conditions, with one recording per subject. Available metadata include clinical severity measures, and all recordings were collected in the \textit{on‑medication} state. Data were de‑identified prior to analysis and used in accordance with institutional and ethical guidelines. The dataset is currently maintained as a private research resource, with provisions for future public release under a research‑only license. Requests may be made for obtaining the dataset once the main study completes.

\subsection{Topological Channel Alignment}
\label{M:Harmonization}

To support cross‑dataset evaluation while preserving electrode geometry, we apply a topological channel alignment based on a unified 65‑channel EEG montage derived from the standard 10–20 system, including the inferior occipital electrode Iz. This reference montage defines the channel set $\mathcal{C}$ serving as a spatial coordinate system across datasets.

Electrodes are grouped into anatomically contiguous regions and ordered according to standard EEG conventions,
\begin{equation}
c \in \mathcal{C}_r =
\begin{cases}
\text{left hemisphere}, & \text{if index is odd}, \\
\text{midline}, & \text{if label ends in $z$}, \\
\text{right hemisphere}, & \text{if index is even}.
\end{cases}
\end{equation}
Each channel is assigned a unique index via a mapping $\pi: \mathcal{C} \rightarrow \{0,1,\dots,C-1\}$, giving a consistent left‑to‑right, anterior‑to‑posterior ordering that is shared across datasets and experimental conditions.

For recordings with incomplete electrode coverage, channels absent from a given dataset are zero‑padded at corresponding indices in the reference. This produces fixed‑size inputs $\mathbf{X} \in \mathbb{R}^{C \times T}$, as illustrated in Fig.~\ref{fig:mp} under \emph{EEG Signals}, explicitly preserving the physical absence of information. No interpolation or imputation is performed, ensuring that spatial structure reflects real‑world acquisition constraints and not learned or inferred channel content.

Importantly, higher‑density EEG layouts (e.g., 10–10 or 10–5 systems) are backward compatible with the standard 10–20 system, such that canonical electrode locations (e.g., C4) remain fixed and are not displaced by the inclusion of additional intermediate sites. In these configurations, higher‑density electrodes are interpolated between established 10–20 landmarks and assigned distinct labels without modifying the original spatial references. Accordingly, restricting analysis to the 10–20 montage (with inclusion of Iz when available) preserves anatomical consistency while avoiding reliance on higher‑density extensions that were not ubiquitously present across datasets.

\subsection{Time‑Frequency Windowing and Representation Variants}
\label{M:TF}

\begin{figure*}[h]
    \centering
    \includegraphics[width=1.0\linewidth]{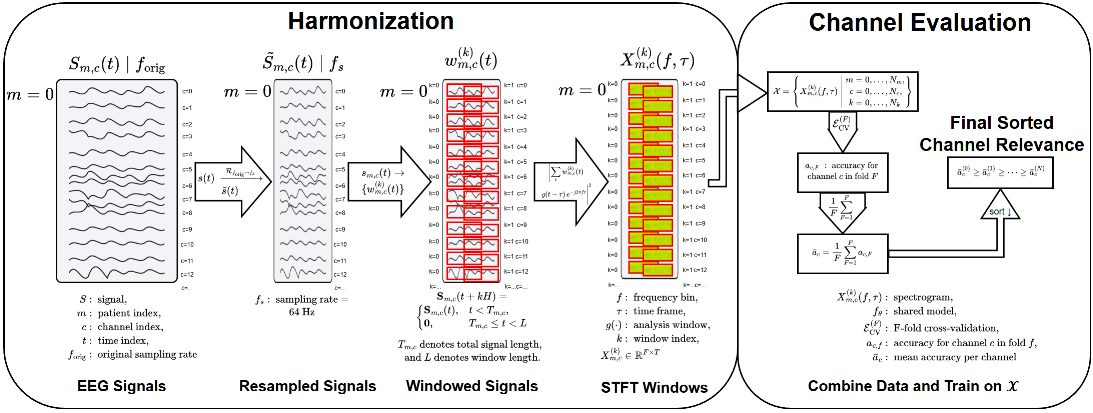}
    \caption{Mathematical pipeline for channel selection. Raw multichannel EEG signals are first preprocessed using standard digital signal processing steps and aligned to target frequency bands. The signals are then segmented into temporal windows and transformed into log‑power spectrograms. A shared model is trained under five‑fold cross‑validation, and channel significance is estimated from validation‑level performance to support population‑aware channel selection.}
    \label{fig:mp}
\end{figure*}

All EEG signals undergo minimal, domain‑agnostic preprocessing prior to time‑frequency analysis. Raw EEG traces $\mathbf{s}_m(t)$ for subject $p_m$, originally sampled at frequency $f_{\text{orig}}$, are resampled to a common rate of $f_s = 64~\text{Hz}$, as illustrated in Fig.~\ref{fig:mp} under \emph{Resampled Signals}. Each resampled signal $\tilde{\mathbf{s}}_m(t)$ is standardized to zero mean and unit variance,

\begin{equation}
\hat{\mathbf{s}}_m(t) = \frac{\tilde{\mathbf{s}}_m(t) - \mu_m}{\sigma_m},
\end{equation}

where $\mu_m$ and $\sigma_m$ denote the mean and standard deviation of $\tilde{\mathbf{s}}_m(t)$. No dataset‑level normalization, band‑pass filtering, or frequency truncation is applied, allowing spectral structure to be learned directly from the data.

Following preprocessing, EEG recordings are partitioned into temporal windows prior to short‑time Fourier transform (STFT) computation, as shown in Fig.~\ref{fig:mp} under \emph{Windowed Signals}. Prior spectrogram‑based analyses have shown that short windows truncate positive features, whereas excessively long windows smear such events across a broader temporal context, reducing discriminability \cite{rasmussen2024deepwhalenet}. Thus, each channel signal is segmented into windows of length $L = 16,384$ samples, corresponding to approximately $256~\text{s}$. Signals shorter than $L$ are zero‑padded according to
\begin{equation}
\mathbf{w}_{m,c}^{(1)}(t) =
\begin{cases}
\hat{\mathbf{s}}_{m,c}(t), & t < T_{m,c}, \\
0, & T_{m,c} \leq t < L,
\end{cases}
\end{equation}
where $T_{m,c}$ denotes the original signal length. For longer recordings, overlapping window use a hop size of $H = L/2$, \[
\mathbf{w}_{m,c}^{(k)}(t) = \hat{\mathbf{s}}_{m,c}(t + kH), \quad k = 0, 1, \dots, K_{m,c},
\] with an additional terminal window included when the final segment is not aligned to the hop size. This window length reflects a balance between temporal context and the preservation of transient discriminative events.

Each window $\mathbf{w}_{m,c}^{(k)}(t)$ is transformed independently using the STFT,
\begin{equation}
\mathbf{X}_{m,c}^{(k)}(f,\tau) =
\left| \sum_{t} \mathbf{w}_{m,c}^{(k)}(t)\, g(t - \tau)\, e^{-j2\pi ft} \right|^2,
\end{equation}
where $g(\cdot)$ denotes the analysis window. For windows of length $L = 16,384$, STFT parameters are fixed to $n_{\text{fft}} = 256$ and hop length $h = 64$. Each resulting spectrogram $\mathbf{X}_{m,c}^{(k)} \in \mathbb{R}^{F \times T}$ is retained at its native resolution and reshaped to $(128, 256, 1)$, as shown in Fig.~\ref{fig:mp} under \emph{STFT Windows}.

Spectrogram magnitudes are normalized on a per‑window basis by their maximum value,

\begin{equation}
\tilde{\mathbf{X}}_{m,c}^{(k)} =
\frac{\mathbf{X}_{m,c}^{(k)}}{\max_{f,\tau} \mathbf{X}_{m,c}^{(k)}},
\end{equation}

and converted to a logarithmic scale. A fixed offset is applied prior to normalization,

\begin{equation}
\hat{\mathbf{X}}_{m,c}^{(k)} =
\frac{\log(\tilde{\mathbf{X}}_{m,c}^{(k)}) - \alpha}
{\max \left( \log(\tilde{\mathbf{X}}_{m,c}^{(k)}) - \alpha \right)},
\quad
\alpha = \min \left( \log(\tilde{\mathbf{X}}_{m,c}^{(k)}) \right),
\end{equation}

after which the first frequency and time bins are removed to mitigate boundary effects introduced by the STFT.

\subsection{Nested Cross‑Validation and Patient‑Level Partitioning}
\label{M:CV}

Within each outer evaluation, a 5-fold cross-validation inner loop schema is used for channel evaluation, as illustrated in Fig.~\ref{fig:mp}. All active channels $\{c_1,\dots,c_C\}$ are evaluated independently, producing validation accuracies $a_{c,f}$ across inner folds $F \in \{1,\dots,N_F\}$. These accuracies are aggregated per channel as
\begin{equation}
\bar{a}_c = \frac{1}{N_F} \sum_{F=1}^{N_F} a_{c,F}.
\end{equation}
Channels are ranked according to $\bar{a}_c$, and the top-ranked subset $\mathcal{C}^\ast$ is retained for evaluation in the corresponding outer fold. Thus, channel selection is driven exclusively by inner-loop evidence and remains isolated from outer-fold test data.

\begin{figure*}[h]
    \centering
    \includegraphics[width=1.0\textwidth]{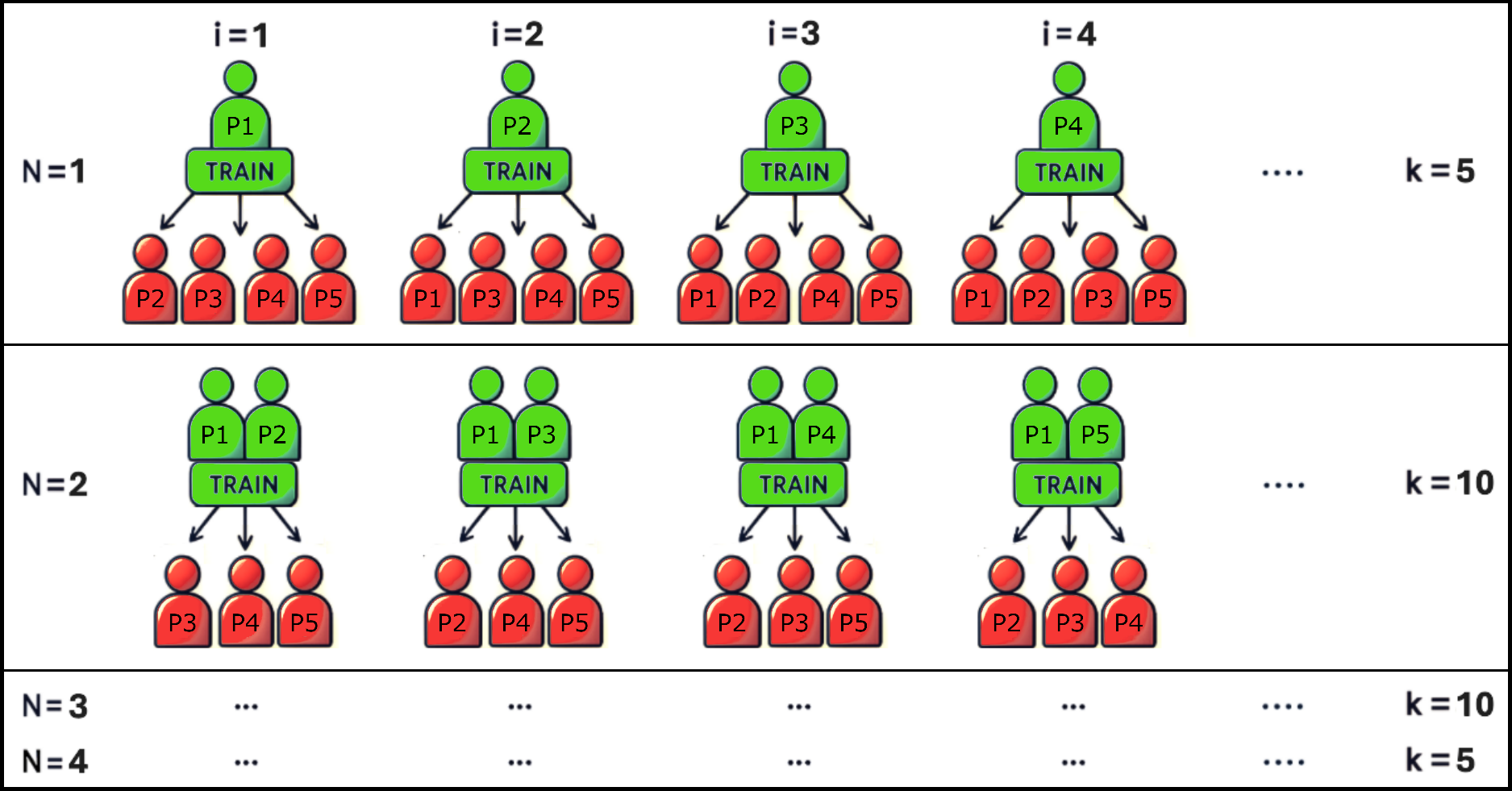}
    \caption{Engineering diagram illustrating structured cross-population evaluation via n-gram expansion. Each row corresponds to a distinct training population size \(n\), with multiple iterations \(i = 1, \ldots, k\) representing different combinations of training and test cohorts. Thus, increasing \(n\) alters the combinatorial structure of population splits, enabling systematic evaluation of generalization across population sizes and configurations. In total, there are 75 outer validations.}
    \label{fig:population_nGram}
\end{figure*}

As shown in Fig.~\ref{fig:population_nGram}, the outer-loop evaluations induced by the cross-dataset $n$-gram design are enumerated as follows. Let $D$ denote the total number of available datasets. The number of $n$-gram levels is $D-1$, corresponding to all possible training-set cardinalates excluding the all-dataset case. For a given level $n \in \{1,\dots,D-1\}$, the number of evaluations is

\begin{equation}
\binom{D}{n} \cdot (D-n),
\end{equation}

where $\binom{D}{n}$ enumerates all selections of $n$ training datasets and $(D-n)$ accounts for each remaining dataset serving as an independent held-out test set. The total number of outer evaluations is obtained by summing across all $n$-gram levels. Model architecture is held constant throughout, and all cross-validation and cross-dataset evaluations are executed using fixed random seeds ($=10$), with inner-loop seeds incremented deterministically to ensure reproducibility. This enumeration strategy is domain-general: it applies to any multi-site evaluation in which population identity must be preserved, independent of the signal modality or classification task.

\subsection{Model Architecture}
\label{M:Model}

We employ a CNN operating on channel‑wise time–frequency representations, with each EEG channel processed independently. This design choice avoids imposing explicit spatial priors across electrodes and allows channel relevance to emerge through data‑driven evaluation. Inputs consist of single‑channel spectrograms $\mathbf{X}_{m,c}^{(k)} \in \mathbb{R}^{F \times T \times 1}$ generated as described in Section~\ref{M:TF}, where each spectrogram corresponds to a single EEG channel and temporal window.

The network architecture comprises a sequence of convolutional blocks, each followed by batch normalization and dropout to promote stable training and mitigate overfitting. Adaptive pooling layers are incorporated to accommodate differences in time–frequency resolution arising from distinct windowing and STFT configurations \cite{rasmussen2025ecologically}. For temporal windows of size $L = 16,384$ samples, a $(2 \times 4)$ pooling operation is applied in the first two convolutional blocks, yielding intermediate feature maps of $64 \times 64$. Subsequent layers progressively increase the number of convolutional filters while reducing spatial resolution, enabling the network to capture increasingly abstract spectral–temporal patterns.

The resulting frequency granularity aligns with canonical EEG rhythms—Delta (0–4~Hz), Theta (4–8~Hz), Alpha (8–12~Hz), and subdivided Beta (12–30~Hz)—without enforcing explicit band boundaries. This alignment facilitates band‑level interpretability of learned features while preserving flexibility in how spectral information is represented. Importantly, the same network architecture is used across all datasets and experimental conditions, ensuring that observed performance differences arise from data characteristics rather than architectural variation.

\subsection{Training Procedure}
\label{M:Training}

Models are trained using stochastic gradient descent with a fixed learning rate of $0.01$ and an exponential decay factor of $0.5$ applied every 10 epochs, for a total of 30 epochs. This schedule provided stable convergence across all evaluated parameterizations. Training is performed at the frame level, with each spectrogram window treated as an independent sample during backpropagation, while model selection and checkpointing are governed exclusively by patient‑level performance as defined in the subsequent paragraph. At the end of each epoch, frame‑level validation predictions are aggregated to produce patient‑level estimates using the evaluation procedure described in the paragraph below. Checkpoints are selected based on patient‑level validation accuracy under a fixed selected‑channel inference regime (top‑$K$, $K=4$), ensuring that retained weights reflect stable behavior under low‑dimensional inference. The same trained weights are subsequently evaluated under alternative inference regimes, including inference using all available channels and reduced subsets ($K=2$ and $K=8$), to assess robustness to channel dimensionality under a fixed optimization regime. Shorter temporal context windows ($L = 2,048$ samples), similar to other studies, were evaluated in preliminary experiments but exhibited substantial run‑to‑run variability under identical training conditions, particularly following the introduction of the expanded 65‑channel montage. Given the focus on robustness and reproducibility under heterogeneous acquisition conditions, these configurations were excluded from the final analysis. All reported experiments therefore use longer temporal contexts ($L = 16,384$), which demonstrated stable behavior across runs and consistent performance.

Although model optimization is performed at the frame level using individual spectrogram windows, all reported performance metrics are computed at the patient level to reflect the intended clinical inference setting. For a given patient $p_m$, frame‑level predicted probabilities $\{\hat{y}_i\}_{i=1}^{n_m}$ spanning all channels and temporal windows are aggregated via soft voting by averaging and thresholding set at $\tau = 0.5$, \[
\hat{Y}_m = \mathbb{I}\!\left[\frac{1}{n_m}\sum_{i=1}^{n_m} \hat{y}_i \ge \tau \right].
\] this aggregation procedure ensures that patient‑level predictions reflect consensus across temporal context and channel observations, not isolated frame‑level decisions. Patient‑level accuracy is computed directly from these aggregated predictions and serves as the primary metric for model selection and comparison. Patient‑level recall and precision are additionally reported to characterize error trade‑offs, from which F1 scores can be inferred forming a complete view of model performance.

\subsection{Population‑Level Analysis}
\label{M:ScalingMetrics}

All reported results are derived from patient‑level predictions aggregated across cross‑validation folds. Cross‑population generalization is analyzed using directional $(\text{train population}, \text{test population})$ pairings, allowing performance to be evaluated asymmetrically depending on the source and target populations. For each ordered population pair, performance metrics are averaged across folds within each channel subset configuration. These directed estimates are organized into square performance matrices, with rows corresponding to training populations and columns to test populations, enabling systematic comparison of cross‑population transfer behavior.

To characterize how generalization performance scales with increasing data diversity, learning‑curve analyses are performed as a function of training population size. For each experiment, effective training population size is defined as

\begin{equation}
N_{\text{train}}(\mathcal{D}_{\text{train}}) = \sum_{d \in \mathcal{D}_{\text{train}}} N(d),
\end{equation}

where $N(d)$ denotes the number of patients in dataset $d$. Patient‑level accuracy is analyzed as a function of $N_{\text{train}}$, pooling results across held‑out test populations and cross‑validation folds to obtain population‑level trends.

Ordinary least squares regression is fit separately for baseline and selected‑channel inference to quantify accuracy scaling with training population size, with confidence intervals computed from the fitted models. All analyses are performed independently for each channel configuration, ensuring that observed scaling behavior reflects population composition, not differences in channel composition.

\subsection{Population‑Specific Channel Selection Maps}
\label{M:ChannelMaps}

Population‑specific EEG channel selection patterns are summarized using three complementary scalp maps: \emph{solo‑trained}, \emph{mixed‑trained}, and \emph{difference} maps. All maps are derived from inner‑loop channel selection outcome calculations and projected onto a unified 65‑channel EEG montage, thus enabling direct comparison across populations under consistent electrode geometry and visualization settings for knowledge extraction.

For each population $p$, solo‑trained maps aggregate channel selection counts from experiments in which the inner‑loop training set consists exclusively of population $p$. For each trained model, the indices of the top‑$4$ selected channels are recorded, and selection frequencies are accumulated across all cross‑validation folds and experimental configurations satisfying this condition. Let $n_{p,c}^{\text{solo}}$ denote the total number of times channel $c$ is selected under solo‑population training. The resulting count vector
\(
\mathbf{n}_p^{\text{solo}} = \{ n_{p,c}^{\text{solo}} \}_{c=1}^{C}
\)
is normalized using min–max scaling,
\(
\hat{\mathbf{n}}_p^{\text{solo}} =
\frac{\mathbf{n}_p^{\text{solo}} - \min_c n_{p,c}^{\text{solo}}}
{\max_c n_{p,c}^{\text{solo}} - \min_c n_{p,c}^{\text{solo}}},
\)
yielding values in $[0,1]$ that are projected onto the standardized scalp layout.

Mixed‑trained maps are constructed by aggregating channel selection counts from all experiments in which population $p$ appears in the inner‑loop training set, regardless of the presence of additional populations. Let $n_{p,c}^{\text{mixed}}$ denote the corresponding selection count for channel $c$, accumulated across all such training mixtures, folds, and configurations. The mixed‑trained count vector \(
\mathbf{n}_p^{\text{mixed}} = \{ n_{p,c}^{\text{mixed}} \}_{c=1}^{C}
\) is normalized using the same min–max procedure, \(
\hat{\mathbf{n}}_p^{\text{mixed}} =
\frac{\mathbf{n}_p^{\text{mixed}} - \min_c n_{p,c}^{\text{mixed}}}
{\max_c n_{p,c}^{\text{mixed}} - \min_c n_{p,c}^{\text{mixed}}},
\) and projected using identical electrode geometry and scaling to compare with solo‑trained maps.

To explicitly quantify differences between population‑specific and mixed‑population channel selection behavior, a difference map is computed for each population $p$ as
\(
\mathbf{d}_p = \hat{\mathbf{n}}_p^{\text{solo}} - \hat{\mathbf{n}}_p^{\text{mixed}}.
\)
Difference values are visualized using a diverging color scale with fixed limits $[-2.5,\,2.5]$, where negative values (blue) indicate channels selected more frequently under mixed‑population training and positive values (red) indicate channels selected more frequently under solo‑population training. All difference maps are rendered using identical limits to facilitating population comparisons.

\section{Empirical Evaluation}
\label{ExpRe}

Our computational framework introduced in Section~\ref{MnM} is evaluated across all n-gram configurations. Thus we report specific dataset characterization, cross-population transfer behavior, and spatial biomarker convergence. We begin by quantifying the composition of the aggregated EEG corpus after preprocessing and windowing, including patient counts, effective signal availability, and sample‑level training exposure across datasets (Section~\ref{DQ}). We next examine the effects of increasing population diversity through the N‑gram training strategy (Section~\ref{nGramEffects}), analyzing how performance and stability scale with training population size (Section~\ref{popScale}), and characterizing directional asymmetries in cross‑population generalization (Section~\ref{subsec:asymmetry}). Finally, we investigate population‑specific and global channel topology patterns to elucidate how spatial evidence is redistributed as training data become more heterogeneous (Section~\ref{popChannelTopo}).

\subsection{Dataset Quantification}
\label{DQ}

To contextualize downstream cross‑population analyses, we first provide a quantitative characterization of the included datasets at both the cohort and signal levels. This subsection summarizes demographic and clinical metadata, effective signal availability, and dataset separability in the learned representation space. Thus, our analyses establish the degree of heterogeneity present across populations and clarify how differences in cohort composition, recording duration, and acquisition context shape the effective training and evaluation landscape.

\paragraph{Cohort Heterogeneity in Demographic and Clinical Metadata}
\label{metaHeterogeneity}

Table \ref{patientMetadata} summarizes the included cohorts and highlights substantial heterogeneity in cohort composition and clinical severity. Sample sizes vary markedly across populations, with PD cohorts ranging from 15 to 82 participants and control cohorts ranging from 16 to 42 participants. Sex distributions also differ across datasets, reflecting variation in recruitment strategies. Despite broadly comparable age distributions across cohorts, clinically meaningful differences are observed in motor symptom severity. In particular, MDS‑UPDRS Part III scores vary substantially across PD populations, with lower average scores observed in the Iowa cohort (P2) and progressively higher scores in the UNM (P1) and OHSU (P4) cohorts. These differences exceed within‑cohort variability and indicate that the datasets sample distinct clinical regimes within the mild‑to‑moderate disease spectrum. In contrast, Hoehn and Yahr staging remains relatively constrained across cohorts, with most PD populations centered around stages 2–3. This suggests that while disease stage inclusion criteria are broadly consistent, motor symptom expression within these stages varies considerably across datasets. Medication state at first acquisition further differs across cohorts, with some datasets including both on‑ and off‑medication recordings and others restricted to on‑medication sessions, contributing additional procedural heterogeneity.

\paragraph{Signal Availability}

Table~\ref{tab:dataset_quantification} summarizes the composition of the aggregated EEG corpus after preprocessing and windowing. Prior to windowing, recordings with less than \textit{30 seconds} of usable signal or with ambiguous or inconsistent naming conventions were excluded to ensure reliable subject identification and sufficient temporal coverage, preventing low‑quality or poorly documented recordings from influencing outcomes. In total, the combined dataset comprises 285 patients drawn from five independent cohorts, yielding 18{,}092 spectrotemporal samples following channel‑wise windowing and STFT transformation. While patient counts vary substantially across datasets, post‑windowing sample contributions are not proportional to cohort size, reflecting differences in recording duration and window yield.

Dataset P2 contributes the largest fraction of training samples, accounting for 44.0\% of all generated samples despite representing approximately one third of the total patients. In contrast, Dataset P3, while containing 31 patients, contributes only 5.7\% of the total samples due to shorter recordings and a reduced channel configuration, thus revealing the structure of acquisition variablility across sites. Windowing effects are further characterized by the distribution of samples per patient. Mean samples per patient range from 33.0 in Dataset~P3 to 100.1 in Dataset~P5, with Dataset~P4 exhibiting a fixed number of samples per subject due to uniform recording length. Concerning raw data availability, recording length is observed directly from the original EEG recordings. Median recording durations range from 2.48 to 5.16 minutes per patient across datasets, with Dataset~P4 providing the longest median coverage. Deriving the raw signal length independent of window overlap and channel count, thus, provides a direct measure of signal availability per subject. We futher report class balance at patient and sample levels to distinguish cohort composition. Patient‑level label counts reflect the underlying distribution of diagnostic classes, while post‑windowing label balance captures model training dynamics. Some datasets exhibit near‑balanced class representations at the patient level and sample‑level. However, training exposure is often more skewed due to differences in patient recording composition.

\paragraph{Dataset Separability \& Embedding Analysis}

\begin{figure*}[h]
    \centering
    \includegraphics[width=1.0\linewidth]{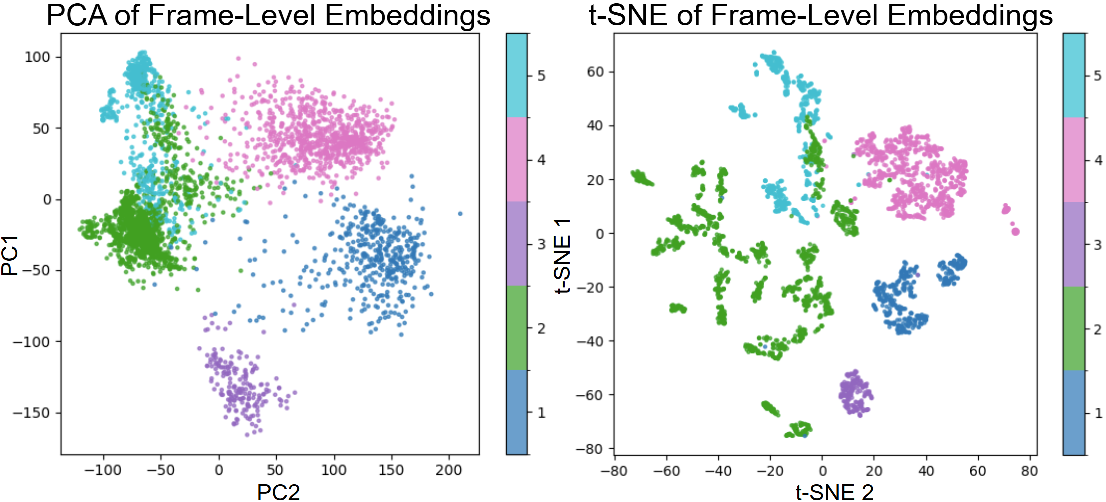}
    \caption{Low‑dimensional projections of frame‑level embeddings for multi‑class dataset identification. PCA (left) and t‑SNE; (right) are computed from the same embedding space and shown for the cross‑validation fold achieving the highest classification accuracy. Each class corresponds to a distinct dataset. Both projections illustrate strong dataset‑specific clustering, with consistent structure observed across all folds in terms of both visualization and quantitative performance.}
    \label{fig:sep}
\end{figure*}

To assess separability between the aggregated datasets prior to downstream classification, we examined low‑dimensional projections of the learned frame‑level embeddings, enforcing patient-level separation between train and test sets to eliminate temporal data leakage. Each dataset was treated as a distinct class, and standard 5‑fold cross‑validation was employed to evaluate consistency across splits. Figure~\ref{fig:sep} presents principal component analysis (PCA) and t‑distributed stochastic neighbor embedding (t‑SNE) visualizations computed from the same embedding space of the fold achieving the highest classification accuracy; all folds exhibited comparable performance and structure. The PCA projection reveals clear separation, indicating that dataset‑specific characteristics are captured in the learned representation without reliance on nonlinear transformation. On the other hand, the t‑SNE visualizations produce compact, well‑defined clusters with minimal overlap between datasets. Importantly, the qualitative organization observed in both projections was consistent across folds, suggesting separability is not driven by a particular data split or visualization method. Supporting these results, quantitative cross‑validation shows most folds achieving classification accuracies exceeding 90\% and area under the ROC curve (AUC) values greater than 97\%. Taken together, the results indicate that the learned embeddings encode robust, low‑noise dataset‑specific structure at the frame level providing empirical confirmation of the computational validity gap identified in Section~\ref{intro}: the learned representation space contains sufficient site-specific structure for a classifier to distinguish acquisition context from disease status, underscoring the necessity of evaluation frameworks that explicitly disentangle sources of variation.

\subsection{N‑Gram Effect Across Configurations}
\label{nGramEffects}

Table~\ref{tab:ab} summarizes classification performance across n‑gram levels, channel configurations, and datasets. These results characterize the behavior of the n-gram evaluation framework introduced in Section~\ref{M:ScalingMetrics}, demonstrating mean performance generally increases with n‑gram length, with consistent improvements observed for n‑gram levels 1–3 and for most datasets at n‑gram~4. Isolated deviations appear at n‑gram~4 corresponding to single‑run evaluations without accompanying variance estimates. In addition to improving mean performance, increases in n‑gram staging is associated with increased stability, reflected by reduced sensitivity to and narrowing performance differences across channel configurations. Furthermore, we can see the effect of channel configuration varies across populations. In P3 and P4, two‑ and four‑channel configurations outperform the baseline at moderate to high n‑gram levels, indicating that targeted channel selection can improve performance when sufficient training diversity is available and the target population is aligned with the training data. Opposingly for P2 and P5, baseline performance frequently matches or exceeds that of selected‑channel configurations, suggesting that channel selection does not always confer benefit across datasets due to channel-wise variability, thus confirming our theoretical stance in Section~\ref{sec:theory_cross_population}. Across all populations, four‑channel models most consistently balance performance and stability, frequently matching or exceeding baseline accuracy while exhibiting lower variability than eight‑channel configurations, which rarely provide additional benefit and occasionally reintroduce the variability seen in the baseline models.

\subsection{Metric Scaling and Stability with Training Population Size}
\label{popScale}

\begin{figure*}[h]
    \centering
    \includegraphics[width=0.90\textwidth]{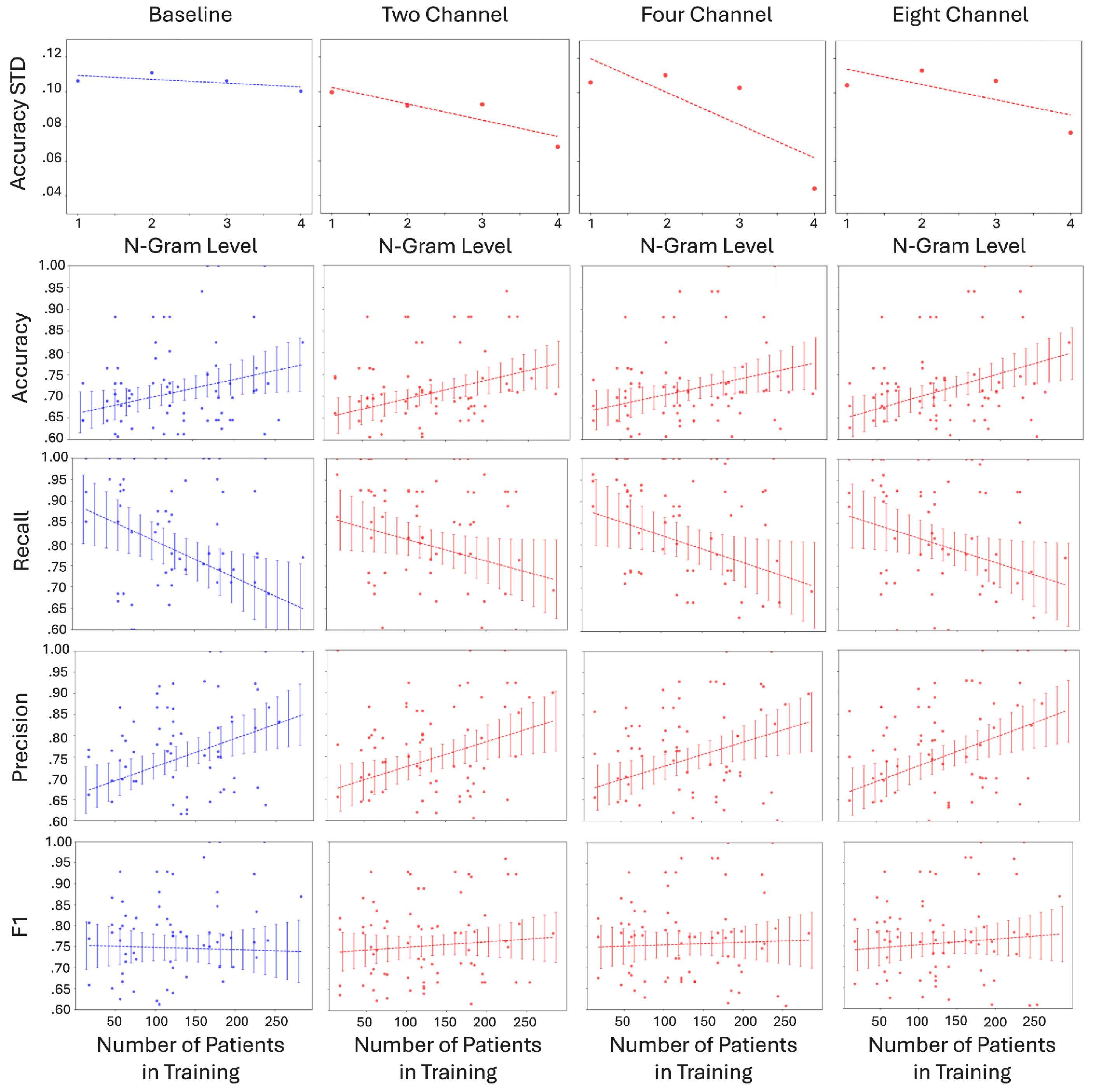}
    \caption{Performance trends across model configurations and training diversity. Columns correspond to the baseline configuration and reduced‑channel models using two, four, and eight channels, respectively. The top row (Row~0) reports the standard deviation of accuracy across all cross‑population tests as a function of N‑gram level, reflecting performance stability under increasing population diversity. Rows~1–4 report accuracy, recall, precision, and F1 score, respectively, plotted against the number of patients included in training across all test populations. All plots include least‑squares regression lines to highlight overall performance trends.}
    \label{fig:population_accuracies}
\end{figure*}

Figure~\ref{fig:population_accuracies} summarizes model performance as a function of training population size across channel configurations. Accuracy variability, measured as the standard deviation across $n$‑gram levels, decreases systematically with increasing training diversity for all configurations, indicating reduced sensitivity to specific population compositions. This stabilization emerges early and is most pronounced for the four‑channel model, which serves as the training objective and exhibits the lowest variability at $n$‑gram level~4. Across all configurations, accuracy increases monotonically with the number of patients included in training, demonstrating that population scaling remains a dominant driver of performance. These gains accrue gradually rather than exhibiting sharp transitions, consistent with progressive alignment between training and test distributions as population diversity increases.

On the other hand, recall and precision exhibit opposing trends as training population size increases, reflecting a redistribution of error types as training diversity grows. Lower‑dimensional channel subsets, particularly the two‑channel configuration, show the least pronounced decline in recall, whereas higher‑dimensional configurations achieve greater precision at larger training sizes. Despite opposing recall and precision dynamics, F1 scores exhibit a modest upward trend across all reduced‑channel configurations, indicating a net improvement in harmonic performance as training population diversity increases. Interestingly, the baseline configuration shows a slight decline in F1 at larger training sizes, suggesting that increased population diversity primarily improves error balance. Notably, trends are metric‑dependent: accuracy driven optimization would be expected to reinforce more conservative decisions. In settings where PD cases outnumber healthy controls, optimizing F1, particularly with respect to the negative class, may therefore be preferable, as it explicitly penalizes asymmetric error accumulation that accuracy alone may obscure.

\subsection{Directional Asymmetry in Cross‑Population Generalization}
\label{subsec:asymmetry}

Table~\ref{tab:asym} reports directional cross‑population classification accuracies under baseline and reduced‑channel configurations. For many dataset pairs, training on population $A$ and testing on population $B$ yields materially different performance than the reverse direction, indicating structured asymmetries in cross‑population generalization. P5 consistently acts as an easy target domain, with models trained on other populations achieving high accuracy when evaluated on P5, while models trained on P5 generalize less reliably to other populations. However, we can see P3 exhibits the opposite pattern: models trained on P3 generalize well across multiple target populations, whereas models trained elsewhere underperform when evaluated on P3. P2 and P4 occupy an intermediate position, exhibiting comparatively balanced cross‑population performance with fewer extreme asymmetries, suggesting more stable biomarker structure across these cohorts. Channel selection modulates absolute performance but does not alter the directional structure of generalization. Directional patterns persist across baseline and reduced‑channel configurations, showing generalization is directional and dataset‑specific, consistent with stable cross‑population biomarker structure.

\subsection{Population‑Specific and Global Channel Topology}
\label{popChannelTopo}

\begin{figure*}[h]
    \centering
    \includegraphics[width=0.95\textwidth]{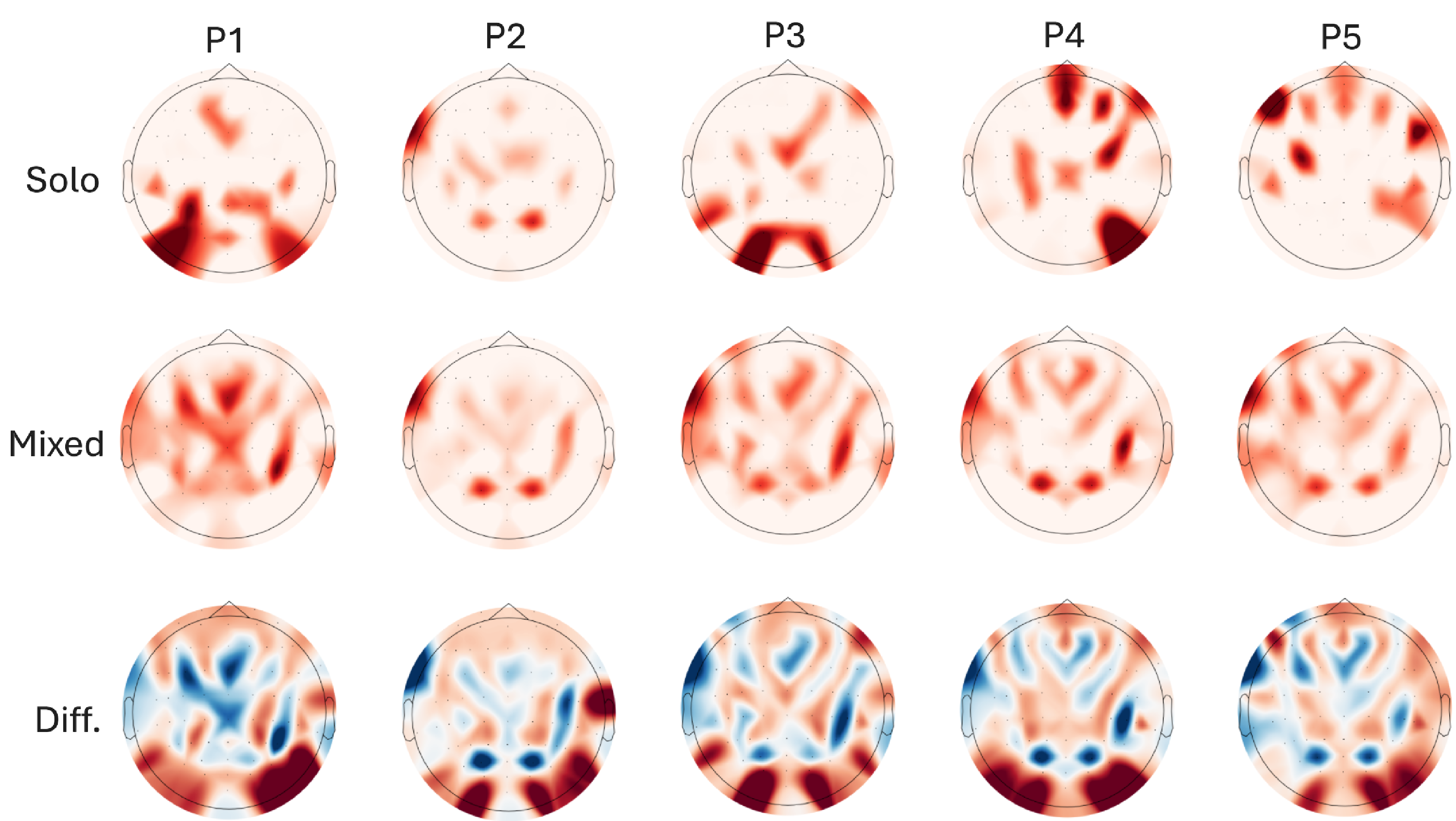}
    \caption{Population‑wise topological EEG channel selection frequency maps derived from inner‑loop channel selection outcomes. Each subfigure corresponds to a distinct population, with channels projected onto a unified 65‑channel EEG montage using topology‑consistent alignment. Color intensity indicates normalized selection frequency, with brighter colors denoting channels selected more frequently across cross‑validation folds and training configurations.}
    \label{fig:population_channel_selection}
\end{figure*}

Figure \ref{fig:population_channel_selection} shows scalp topographies of channel selection frequency under single‑population training, mixed‑population training, and their difference. Single‑population maps exhibit mild population‑specific variation but converge on common spatial patterns, with frequent selections along the posterior occipital–cervical boundary, frontal crown, and bilateral temporal regions. This convergence indicates shared structure across independently trained models despite population‑specific differences and reflects stable underlying biomarker organization. Mixed‑population training produces more spatially coherent and less fragmented channel topologies, consistent with reinforcement of features that generalize across datasets. Difference maps make this transition explicit by separating population‑dependent channels emphasized under single‑population training from regions reinforced under mixed training. As training population size and sample count increase, channel importance increasingly aggregates toward patterns observed in Dataset~P2, reflecting its larger patient cohort and greater contribution to the training distribution. Notably, many single‑population maps exhibit activation around these same P2‑associated nodes, suggesting that the spatial patterns emphasized in P2 capture disease‑relevant structure that generalizes across populations. Thus, increased training diversity redistributes channel importance toward globally consistent, disease‑informative structure while not amplifying population‑specific patterns. Thus, the convergence of channel importance under mixed-population training shows empirical evidence for the geometric projection interpretation of channel selection developed in Section~\ref{sec:theory_cross_population}.


\section{Discussion}
\label{S:Discussion}

Three lenses dominate the discussion: empirical characterization of cross-population biomarker transfer dynamics (Section~\ref{subsec:empirical_discussion}), formal mixture-risk analysis grounding the observed behaviors in hypothesis-space geometry (Section~\ref{sec:theory_cross_population}), and derivation of design principles for cross-population evaluation in multi-site biomedical signal classification (Section~\ref{IEBCM}). The empirical findings map onto the four contributions declared in Section~1: directional and non-reciprocal biomarker transfer across populations (Contribution~1), population-scaling behavior under the n-gram evaluation framework (Contribution~2), a formal mixture-risk analysis revealing structural consequences of multi-population training geometry (Contribution~3), and a population-aware channel-knowledge extraction procedure for robust spatial biomarkers that characterizes channel selection as a geometric projection (Contribution~4).

\subsection{Empirical Analysis of Framework Behavior}
\label{subsec:empirical_discussion}

\paragraph{Representational separability validates the computational validity gap.}

Dataset‑separability analysis in Section~\ref{DQ} gives direct empirical evidence for the computational validity gap identified in Section~\ref{intro}. Representation learning in this setting is not subtle. The learned embedding space encodes dataset identity with classification accuracy above 90\% and ROC‑AUC exceeding 97\%, showing that site‑specific acquisition structure is sufficiently prominent for a simple classifier to recover cohort provenance with high reliability. Linear (PCA) and nonlinear (t‑SNE) projections both yield compact, well‑separated clusters by site, which rules out the possibility that the observed structure is an artifact of a particular visualization method or data split. These patterns support the view that models trained on heterogeneous multi‑site EEG can internalize cohort‑specific structure, hardware fingerprints, preprocessing artifacts, or demographic confounds that overshadow disease‑relevant neurophysiology \cite{geirhos2020shortcut, zech2018variable}. Taken together, the separability results establish the empirical backdrop for the analyses that follow, which examine how directional biomarker extraction behaves when the underlying representation space is already stratified by acquisition site.

\paragraph{Directional and non-reciprocal biomarker transfer.}

Within‑dataset cross‑validation remains strong across all populations, a pattern that makes the subsequent cross‑population failures more striking. Models that appear stable under i.i.d. evaluation often collapse when tested on an unseen cohort, revealing a sharp divergence between within‑site performance and genuine out‑of‑distribution behavior. Earlier EEG studies have reported the same tension between high subject‑ or dataset accuracy and poor external generalization \cite{roy2019deep, schirrmeister2017deep, rasmussen2025channel}, with the present results following the same trajectory in an explicit population setting. Against this backdrop, the directional nature of cross‑population transfer becomes clearer. Some populations serve as reliable sources while others do not, and the asymmetry persists across channel‑selection strategies. Prior work on dataset bias and non‑reciprocal generalization in high‑dimensional systems describes similar source–target imbalances \cite{torralba2011unbiased}, but the magnitude observed here is unusually pronounced. These directional gaps form the core of Contribution~1, which formalizes how the relationship between source and target distributions governs transfer behavior rather than any single measure of aggregate accuracy. Population‑specific acquisition characteristics therefore exert a dominant influence on biomarkers, even when the underlying representations and model architectures are held fixed \cite{shafiezadeh2023methodological}. The mixture‑risk bound in Section~\ref{sec:theory_cross_population} provides a theoretical framing: transfer gaps arise from distributional discrepancy between the training mixture and the held‑out target, and that discrepancy constrains the reliability of any biomarker extracted from heterogeneous multi‑site EEG.

\paragraph{Population-scaling behavior under the n-gram framework.}

Our study mirrors patterns reported in domain‑generalization studies, where exposure to multiple environments suppresses dataset‑specific biases and encourages more stable representations \cite{gulrajani2020search, torralba2011unbiased}. Population‑level scaling behaves in a surprisingly regular way once training diversity begins to increase. Performance jumps sharply when moving from single‑population to multi‑population training, and the variance across held‑out test populations contracts at the same time. Contribution~2 formalizes this behavior within the n‑gram framework, which elucidates a clean way to track how biomarkers evolves as additional cohorts are introduced. The gains, however, do not scale indefinitely. After a moderate level of population diversity, additional datasets yield only marginal improvements in absolute accuracy. Prior scaling analyses in deep learning describe similar saturation effects, where more data primarily reduces variance and redistributes error while not producing proportional gains in task performance \cite{hestness2017deep, belkin2019reconciling}. Here, that redistribution appears as opposing recall–precision dynamics: recall declines while precision rises, a shift that reflects changes in the model’s error profile as training diversity grows. This pattern aligns with the goal of learning a population‑robust representation rather than maximizing i.i.d. accuracy. A model trained on diverse cohorts serves as a stable initialization that can later be adapted to population‑specific or i.i.d. settings, where additional data may again produce larger gains \cite{pan2009survey, yosinski2014transferable}. Within the cross‑population evaluation protocol, training diversity therefore acts primarily as a stabilizing force under distributional shift, with only modest improvements in absolute performance once heterogeneity is sufficiently broad.

\paragraph{Geometric Convergence of Channel Structure and Sensitivity–Specificity Control.}

Cohort‑dependent spatial structure is a well‑documented feature of multi‑site EEG and M/EEG studies, where acquisition protocols and recording environments shape channel relevance in systematic ways \cite{pernet2020issues}, and this same structure governs how channel dimensionality modulates precision and recall across populations. When models are trained on a single cohort, they emphasize spatial and spectral characteristics unique to that population, a pattern that Contribution~4 captures by treating channel selection as a mechanism for extracting cohort‑level structure; yet these same low‑dimensional subsets tend to preserve sensitivity, revealing that the geometry of channel restriction doubles as an operating‑point control. Mixed‑population training shifts the picture. Channel importance migrates toward regions that remain informative across datasets, producing more coherent and globally stable selection patterns that support robust biomarker extraction, while larger channel sets sharpen specificity by enabling more selective representations that reduce false positives \cite{blankertz2011single, gulrajani2020search}. Difference maps make both transitions visible: channels that dominate under solo training often diminish as additional populations are introduced, and channels that rise in prominence tend to reflect population‑invariant structure, just as expanding the channel set redistributes error and sharpens decision boundaries without uniformly improving all metrics \cite{hestness2017deep, belkin2019reconciling}. This redistribution suggests that the channel‑knowledge extraction procedure suppresses cohort‑specific idiosyncrasies and amplifies shared discriminative structure, while simultaneously tuning the sensitivity–specificity balance through representational capacity. The idea aligns with broader work in invariant representation learning, where exposure to multiple environments reduces reliance on spurious, environment‑dependent features while preserving stable signal components \cite{arjovsky2019invariant}, thus Section~\ref{sec:theory_cross_population} formalizes this intuition through a geometric view of channel selection in which restricting dimensionality collapses population‑specific nuisance directions while retaining shared structure embedded in the learned representation.

\subsection{Formal Theoretical Analysis}
\label{sec:theory_cross_population}

The empirical behaviors described above share a common origin. They emerge from the geometry imposed by multi‑population training, not from isolated quirks of individual datasets. The mixture‑risk perspective developed below (Contribution~3) makes this connection explicit. Using standard notation from domain‑generalization theory \cite{ben2010theory, mansour2008domain}, the formalism isolates the minimal distributional mechanisms capable of producing the observed patterns. Each mechanism yields concrete, testable predictions, several of which appear directly in the evaluation results summarized in Section~\ref{ExpRe}. Let each EEG cohort \(d \in \{1,\dots,D\}\) define a distribution \(\mathcal{P}_d(X,Y)\) over representations \(X\) and labels \(Y \in \{0,1\}\). For a training subset \(S \subseteq \{1,\dots,D\}\), define the mixture distribution \[
\mathcal{P}_S = \sum_{d \in S} \alpha_d \mathcal{P}_d, \qquad \alpha_d \ge 0,\ \sum_{d \in S} \alpha_d = 1,
\] and the population risk \[
R_d(f) = \mathbb{E}_{(X,Y)\sim\mathcal{P}_d}\big[\ell(f(X),Y)\big],
\] with mixture risk \[
R_S(f) = \sum_{d \in S} \alpha_d R_d(f).
\] Empirical risk minimization on pooled data therefore targets \(R_S\), not the risk of any individual population, and this mismatch is the starting point for understanding how robust features behave across heterogeneous cohorts.

\paragraph{Directional transfer as mixture mismatch.}

For a held-out population \(t \notin S\), define the transfer gap

\[
\Delta_{S \to t}(f) = R_t(f) - R_S(f).
\]

A standard domain-generalization bound yields

\[
R_t(f) \;\le\; R_S(f) + \operatorname{disc}(\mathcal{P}_S,\mathcal{P}_t) + \lambda_{S,t},
\]

where \(\operatorname{disc}(\cdot,\cdot)\) measures distributional discrepancy and \(\lambda_{S,t}\) captures irreducible joint error due to label ambiguity or non-shared discriminative structure \cite{ben2010theory, mansour2008domain}. This inequality highlights a structural property of cross-population learning: minimizing \(R_S\) does not guarantee small \(R_t\) unless the target distribution lies near the convex hull of the training mixture. Because \(\mathcal{P}_S\) depends on the specific populations and weights \(\{\alpha_d\}\), cross-population biomarker transfer is inherently directional and non-reciprocal \cite{torralba2011unbiased, gulrajani2020search}. The directional asymmetries documented in Section~\ref{subsec:asymmetry} conform to this prediction: populations that lie farther from the training mixture's convex hull exhibit larger transfer gaps, regardless of channel configuration or other hyperparameter setting that can confound performance estimates.

\paragraph{Hypothesis-space contraction under multi-population training.}

For \(\varepsilon \ge 0\), define the set of \(\varepsilon\)-good classifiers for population \(d\) as \[
\mathcal{H}_d(\varepsilon) = \{f \in \mathcal{F} : R_d(f) \le \varepsilon\}.
\] Training on a population set \(S\) restricts feasible solutions to the intersection \[
\mathcal{H}_S(\varepsilon) = \bigcap_{d \in S} \mathcal{H}_d(\varepsilon).
\] As additional, heterogeneous populations are included, this intersection contracts: \[
\mathcal{H}_{S_2}(\varepsilon) \subseteq \mathcal{H}_{S_1}(\varepsilon), \qquad S_1 \subseteq S_2.
\] Classifiers that rely on population‑specific shortcuts cannot maintain low risk across all domains and are progressively removed, while those supported by structure shared across populations remain viable. This contraction explains reduced performance variance, improved stability under distributional shift, and the saturation of mean gains as training‑population diversity increases \cite{belkin2019reconciling, hestness2017deep, gulrajani2020search}. It also aligns with invariant‑representation perspectives without requiring an explicit decomposition of the input space \cite{arjovsky2019invariant}. The diminishing‑returns behavior observed across n‑gram levels in Section~\ref{popScale} follows directly: once population‑specific shortcuts have been pruned from the feasible set, additional cohorts yield progressively smaller reductions in hypothesis‑space volume.

\paragraph{Channel selection as geometric projection.}

Let \(C\) denote the full channel set and \(S_C \subseteq C\) a selected subset. Channel selection induces a projection of the representation space and an associated restricted hypothesis class \[
\mathcal{F}_{S_C} = \{f \in \mathcal{F} : f \text{ depends only on channels in } S_C\}.
\] This restriction modifies both model complexity and the geometry of the induced distributions \(\mathcal{P}_d^{S_C}\). A concise risk bound after channel restriction is \[
R_t(f_{S_C}) \le \hat{R}_S(f_{S_C}) + \mathrm{Complexity}(\mathcal{F}_{S_C}) + \operatorname{disc}(\mathcal{P}_S^{S_C},\mathcal{P}_t^{S_C}),
\] where \(f_{S_C} \in \mathcal{F}_{S_C}\) and \(\mathcal{P}_S^{S_C}\), \(\mathcal{P}_t^{S_C}\) denote the projected distributions \cite{ben2010theory, guyon2003introduction}. Geometrically, reducing channel dimensionality can collapse population-specific nuisance directions, decreasing discrepancy, while simultaneously attenuating weak but informative disease-related structure. This redistribution of error shifts the model's operating point, yielding systematic precision--recall trade-offs rather than uniform performance gains \cite{blankertz2011single, hestness2017deep}. The monotonic precision--recall separation observed across channel configurations in Section~\ref{popScale} conforms to this geometric interpretation: each reduction in channel count trades discriminative resolution for reduced distributional discrepancy.

\paragraph{Synthesis.}

The three perspectives developed above describe a single underlying geometry that governs cross‑population EEG biomarker behavior. Directional transfer gaps follow directly from distributional mismatch between the training mixture and a held‑out target, a relationship made explicit by the mixture‑risk bound. The same geometry explains why population‑scaling gains saturate: as additional cohorts are introduced, the feasible hypothesis space contracts, pruning population‑specific shortcuts and reducing variance without producing unbounded improvements in mean accuracy \cite{belkin2019reconciling, hestness2017deep, gulrajani2020search}. Channel‑selection dynamics fit naturally into this picture. Restricting the channel set attenuats weak but informative structure, yielding the systematic precision–recall shifts observed across configurations \cite{blankertz2011single, arjovsky2019invariant}. Thus, biomarkers are shaped primarily by population structure and distributional geometry not model capacity, and that the evaluation framework exposes these effects in a reproducible way applicable across signal domains \cite{torralba2011unbiased, arjovsky2019invariant, hestness2017deep, belkin2019reconciling, gulrajani2020search}.

\subsection{Design Principles for Cross-Population Biomedical Signal Evaluation}
\label{IEBCM}

Cross‑population evaluation is shaped less by model choice than by the geometry of the underlying data. The theoretical analysis in Section~\ref{sec:theory_cross_population} make this clear: large directional transfer gaps signal high discrepancy \(\operatorname{disc}(\mathcal{P}_S,\mathcal{P}_t)\), stability gains from multi‑site training reflect contraction of the feasible hypothesis set \(\mathcal{H}_S(\varepsilon)\), and channel restriction acts as a projection that reshapes both the induced distributions \(\mathcal{P}^{S_C}\) and the effective complexity \(\mathrm{Complexity}(\mathcal{F}_{S_C})\). These mechanisms have direct practical consequences. The remainder of this subsection translates them into concrete guidance for evaluation protocol design, data‑collection strategy, representational architecture, and deployment‑time monitoring.

\paragraph{Directional Testing and Cohort‑Diverse Training Mixtures.}

Cross‑population evaluation must respect the asymmetry built into the mixture‑risk setting, because minimizing \(R_S\) says little about \(R_t\) when \(\mathcal{P}_t\) lies outside the training mixture, and this same asymmetry highlights why robustness depends more on who is in the training mixture than on how many samples come from any single site. Preserving train–test directionality through cohort‑level block testing—leave‑one‑cohort‑out or multi‑cohort holdouts—provides a far more realistic view than a single 80/20 split, which assumes homogeneous data and often yields optimistic estimates \cite{torralba2011unbiased, pernet2020issues}, just as adding heterogeneous cohorts removes population‑specific shortcuts and lowers variance across unseen targets. Reports should present directional transfer gaps \(\Delta_{S\to t}\) and variance across held‑out cohorts alongside pooled accuracy \cite{ben2010theory, gulrajani2020search}, because once \(\mathcal{H}_S(\varepsilon)\) stabilizes, additional samples from a single site contribute little, whereas new acquisition conditions or cohort characteristics continue to reshape the feasible set \cite{belkin2019reconciling, hestness2017deep}. This level of reporting clarifies where models succeed—and where they fail—across populations, while transparent documentation of mixture weights \(\{\alpha_d\}\) allows downstream users to assess representational coverage and anticipate robustness under distributional shift. In practice, these considerations jointly motivate multi‑site consortia, targeted collection of underrepresented modalities, and evaluation protocols that expose stability rather than mask it, ensuring that empirical performance reflects population‑level generalization rather than idiosyncrasies of any single cohort helping to transfer the learned representations.

\paragraph{Channel selection and representational trade-offs.}

Channel selection is a geometric choice, not a computational afterthought. Reducing the channel set projects each population into a lower‑dimensional subspace, which can shrink \(\operatorname{disc}(\mathcal{P}_S^{S_C},\mathcal{P}_t^{S_C})\) by collapsing nuisance directions, but it may also weaken invariant disease‑related structure and increase irreducible error. The operating point therefore has to be chosen deliberately. Lower‑dimensional subsets tend to favor sensitivity and are well suited to screening contexts, whereas larger subsets support more selective, high‑precision decisions in settings where false positives carry higher cost \cite{guyon2003introduction, blankertz2011single}. Empirical selection procedures should report how channel choices alter both \(\mathrm{Complexity}(\mathcal{F}_{S_C})\) and cross‑population discrepancy, making the resulting operating regime transparent to downstream users.

\paragraph{Spectral–Temporal Robustness and Deployment‑Stage Drift Control.}

Overly rigid feature sets—such as relying on a single electrode’s band power—tend to fracture under population shift because they exclude invariant structure that persists across cohorts \cite{S0219635225009532}, and this same brittleness becomes a deployment concern when target populations fall outside the convex hull of the original training mixture. A data‑driven spectral–temporal representation avoids this fragility by allowing the model to align naturally with canonical rhythms (Delta, Theta, Alpha, Beta) \cite{schirrmeister2017deep, gabeff2021interpreting}, yet even flexible representations require monitoring mechanisms that track shifts in \(\operatorname{disc}(\mathcal{P}_S,\mathcal{P}_{\text{deploy}})\) and performance on representative holdouts. Given the sampling rate (\(64~\text{Hz}\)), learned features are inherently restricted to sub‑gamma frequencies (\(<32~\text{Hz}\)), which supports interpretable band‑level comparisons and reduces sensitivity to high‑frequency acquisition artifacts, while global models trained on diverse cohorts provide stable initializations for lightweight, site‑specific fine‑tuning when labeled data are available \cite{pan2009survey, yosinski2014transferable}. Presenting band‑level activations alongside channel‑level maps helps clinicians and engineers identify population‑dependent spectral emphasis \cite{rasmussen2025channel, shafiezadeh2023methodological}, and model selection in deployment should reflect the clinical costs of different error types, since representation choices systematically trade sensitivity against specificity. These considerations extend beyond EEG‑based PD detection: any multi‑site biomedical signal system operating under non‑stationary acquisition conditions requires representations that remain interpretable under shift and monitoring procedures that detect representational drift, ensuring reliability in real‑world settings.

\paragraph{Domain generality of the framework and theory.}

Our computational infrastructure is not specific to EEG or to Parkinson’s disease. The n‑gram expansion strategy (Contribution~2) applies whenever data arise from \(D\) identifiable populations and the goal is to characterize how extracted features depend on training‑population composition. Likewise, the mixture‑risk bound, hypothesis‑space contraction, and geometric projection mechanisms (Contribution~3) are stated in terms of generic distributions \(\mathcal{P}_d(X,Y)\) and hypothesis classes \(\mathcal{F}\), with no assumptions tied to EEG structure. The channel‑knowledge extraction procedure (Contribution~4) extends naturally to any modality where spatial or sensor‑level feature selection matters. As a result, the framework generalizes to multi‑site evaluation in other biomedical signal domains (e.g., ECG, EMG, wearable accelerometry), to medical imaging across institutional pipelines, and even to ecological monitoring under heterogeneous acquisition conditions \cite{rasmussen2025ecologically}. The present study validates the approach on a challenging case—low‑SNR, high‑heterogeneity EEG from five independent cohorts—but the design principles and theoretical guarantees are intended as reusable methodology for the broader computing‑for‑healthcare community.

\section{Limitations}
\label{Lim}

Our study is not intended to optimize leaderboard performance or to exhaustively search the space of architectures, training strategies, or evaluation metrics. The computational framework is deliberately constrained so that population diversity, cohort specificity, and data directionality can be examined without interference from model‑level confounders. More elaborate architectures, alternative channel‑selection criteria, multiple‑instance learning formulations, richer feature engineering, or task‑specific metrics could all raise absolute accuracy, but they would obscure the population‑level effects that are central to this work. The datasets themselves introduce another limitation: acquisition protocols and cohort characteristics vary substantially across sites, which likely suppresses peak performance relative to more standardized settings. This heterogeneity is intentional. It functions as a stress test for generalization, and the framework’s ability to scale across diverse populations reflects robustness to real‑world variability rather than reliance on cohort‑specific structure. The reported results should therefore be interpreted as characterizing emergent biomarker‑transfer behavior under population diversity, not as defining upper bounds on achievable performance.

A second limitation concerns the clinical composition of the datasets. All cohorts consist of individuals with established, clinically diagnosed Parkinson’s disease, many of whom are receiving treatment and are several years post‑diagnosis. The models evaluated here should not be interpreted as performing de novo diagnosis or early‑stage screening in untreated or prodromal populations. The task studied is the detection of disease‑associated EEG biomarkers in confirmed PD. Although some cohorts include individuals at relatively early stages, the present results do not address diagnostic sensitivity at initial presentation or differentiation between idiopathic and atypical parkinsonian syndromes. These remain important directions for future work and will require purpose‑built datasets with more granular characterization of biomarker evolution across disease stages.

\paragraph{Computational and theoretical scope.}

The mixture‑risk analysis in Section~\ref{sec:theory_cross_population} (Contribution~3) identifies structural mechanisms that explain the qualitative empirical behaviors observed in this study. These mechanisms are expressed as inequalities and set‑containment relations rather than tight equalities. They clarify \emph{why} directional transfer gaps, scaling saturation, and precision–recall trade‑offs arise, but they do not predict their exact magnitudes for any specific pair of populations. Tightening these bounds would require estimating distributional discrepancy \(\operatorname{disc}(\mathcal{P}_S,\mathcal{P}_t)\) and irreducible joint error \(\lambda_{S,t}\) from finite samples, an open problem in domain‑generalization theory \cite{ben2010theory}. The n‑gram evaluation framework (Contribution~2) also carries computational limits: enumerating all \(\binom{D}{n}\) train–test combinations scales combinatorially with the number of cohorts, and studies with substantially more than five populations may require approximation strategies such as stratified subsampling. Finally, although the framework, theory, and channel‑knowledge extraction procedure (Contribution~4) are stated in domain‑general terms, the empirical validation is restricted to resting‑state EEG classification with a single CNN architecture. Demonstrating generality across signal modalities, task‑state paradigms, and model families remains an important direction for future work, particularly in settings with more complex population structure.

\section{Conclusion}
\label{Conc}

We introduced a computational framework for evaluating cross‑population generalization in multi‑site biomedical signal classification and validated it on five heterogeneous EEG cohorts (285 participants). The study establishes four main contributions: (1) cross‑population biomarker transfer is directional and non‑reciprocal, and symmetric evaluation obscures this structure; (2) an n‑gram evaluation framework enumerates all population‑level train–test configurations, enabling principled analysis of how performance scales with training‑population diversity; (3) a mixture‑risk formulation grounded in hypothesis‑space contraction explains directional transfer gaps, population‑scaling saturation, and precision–recall trade‑offs as structural consequences of multi‑population training geometry; and (4) a population‑aware channel‑knowledge extraction method identifies spatially coherent, population‑invariant biomarker candidates by interpreting channel selection as a geometric projection that trades representational complexity for reduced distributional discrepancy. Our results show that robust biomarker discovery under population heterogeneity requires evaluation protocols that preserve directionality and expose stability, shifting the objective from maximizing within‑cohort accuracy to identifying spatial signatures that remain invariant under distributional shift. The framework and theory, demonstrated here on low‑SNR, high‑heterogeneity EEG, are domain‑general and provide reusable methodology for any setting where biomarker validity must be established across diverse populations.

\section*{Acknowledgments}
\label{Ack}

This work was supported by the National Science Foundation under Grant No.~\href{https://www.nsf.gov/awardsearch/showAward?AWD_ID=2346643}{\#2346643}, the U.S. Department of Defense under Award No.~\href{https://dtic.dimensions.ai/details/grant/grant.14525543}{\#FA9550-23-1-0495}, and the U.S. Department of Education under Grant No.~P116Z240151. Computations supporting this project were performed on High Performance Computing systems at the University of South Dakota, funded by NSF Award OAC‑1626516. This work was supported in part by the Parkinson’s Foundation fellowship for basic scientists to Dr Samuel Stuart (PF-FBS-1898-18-01). 

Any opinions, findings, conclusions, or recommendations expressed in this material are those of the author(s) and do not necessarily reflect the views of the National Science Foundation, the U.S. Department of Defense, or the U.S. Department of Education. The author(s) acknowledges Microsoft Copilot for assistance with grammar, structural edits, contextual extrapolation, and figure preparation, all with extensive human input~\cite{CoPilot}. All core ideas, analyses, and interpretations are the authors' own, based on prior literature and experimental analysis as cited in the manuscript.

{\footnotesize
\bibliographystyle{IEEEtran}
\bibliography{IEEEabrv, eegGen}

\begin{thebibliography}{10}
\providecommand{\url}[1]{#1}
\csname url@samestyle\endcsname
\providecommand{\newblock}{\relax}
\providecommand{\bibinfo}[2]{#2}
\providecommand{\BIBentrySTDinterwordspacing}{\spaceskip=0pt\relax}
\providecommand{\BIBentryALTinterwordstretchfactor}{4}
\providecommand{\BIBentryALTinterwordspacing}{\spaceskip=\fontdimen2\font plus
\BIBentryALTinterwordstretchfactor\fontdimen3\font minus \fontdimen4\font\relax}
\providecommand{\BIBforeignlanguage}[2]{{%
\expandafter\ifx\csname l@#1\endcsname\relax
\typeout{** WARNING: IEEEtran.bst: No hyphenation pattern has been}%
\typeout{** loaded for the language `#1'. Using the pattern for}%
\typeout{** the default language instead.}%
\else
\language=\csname l@#1\endcsname
\fi
#2}}
\providecommand{\BIBdecl}{\relax}
\BIBdecl

\bibitem{shalev2014understanding}
S.~Shalev-Shwartz and S.~Ben-David, \emph{Understanding Machine Learning: From Theory to Algorithms}.\hskip 1em plus 0.5em minus 0.4em\relax Cambridge, UK: Cambridge University Press, 2014.

\bibitem{vapnik1998statistical}
V.~N. Vapnik, \emph{Statistical Learning Theory}.\hskip 1em plus 0.5em minus 0.4em\relax New York: Wiley, 1998.

\bibitem{quinonero2009dataset}
J.~Quionero-Candela, M.~Sugiyama, A.~Schwaighofer, and N.~D. Lawrence, ``Dataset shift in machine learning,'' \emph{MIT Press}, 2009.

\bibitem{bradshaw2023crossvalidation}
\BIBentryALTinterwordspacing
T.~J. Bradshaw, Z.~Huemann, J.~Hu, and A.~Rahmim, ``A guide to cross-validation for artificial intelligence in medical imaging,'' \emph{Radiology: Artificial Intelligence}, vol.~5, no.~4, p. e220232, 2023. [Online]. Available: \url{https://doi.org/10.1148/ryai.220232}
\BIBentrySTDinterwordspacing

\bibitem{geirhos2020shortcut}
R.~Geirhos, J.-H. Jacobsen, C.~Michaelis, R.~Zemel, W.~Brendel, M.~Bethge, and F.~A. Wichmann, ``Shortcut learning in deep neural networks,'' \emph{Nature Machine Intelligence}, vol.~2, no.~11, pp. 665--673, 2020.

\bibitem{zech2018variable}
J.~R. Zech, M.~A. Badgeley, M.~Liu, A.~B. Costa, J.~J. Titano, and E.~K. Oermann, ``Variable generalization performance of a deep learning model to detect pneumonia in chest radiographs: A cross-sectional study,'' \emph{PLOS Medicine}, vol.~15, no.~11, p. e1002683, 2018.

\bibitem{compton2023when}
R.~Compton, L.~Zhang, A.~Puli, and R.~Ranganath, ``When more is less: Incorporating additional datasets can hurt performance by introducing spurious correlations,'' in \emph{Proceedings of Machine Learning Research}, vol. 219, 2023, pp. 1--24.

\bibitem{cross2024bias}
J.~L. Cross, M.~A. Choma, and J.~A. Onofrey, ``Bias in medical ai: Implications for clinical decision-making,'' \emph{PLOS Digital Health}, vol.~3, no.~11, p. e0000651, 2024.

\bibitem{roberts2021common}
M.~Roberts, D.~Driggs, M.~Thorpe, J.~Gilbey, M.~Yeung, S.~Ursprung, A.~I. Aviles-Rivero, C.~Etmann, C.~McCague, L.~Beer \emph{et~al.}, ``Common pitfalls and recommendations for using machine learning to detect and prognosticate for covid-19 using chest radiographs and ct scans,'' \emph{Nature Machine Intelligence}, vol.~3, no.~3, pp. 199--217, 2021.

\bibitem{avola2025benchmarking}
D.~Avola, A.~Bernardini, G.~Crocetti, A.~Ladogana, M.~Lezoche, M.~Mancini, D.~Pannone, and A.~Ranaldi, ``Benchmarking of eeg analysis techniques for parkinson's disease diagnosis: A comparison between traditional ml methods and foundation dl methods,'' \emph{arXiv preprint arXiv:2507.13716}, 2025.

\bibitem{anjum2024resting}
M.~F. Anjum, A.~I. Espinoza, R.~C. Cole, A.~Singh, P.~May, E.~Y. Uc, S.~Dasgupta, and N.~S. Narayanan, ``Resting-state eeg measures cognitive impairment in parkinson’s disease,'' \emph{npj Parkinson’s Disease}, vol.~10, no.~6, 2024.

\bibitem{may2023portlandEEG}
\BIBentryALTinterwordspacing
P.~May, E.~Y. Uc, and N.~S. Narayanan, ``Eeg mortality dataset in parkinson’s disease,'' 2023. [Online]. Available: \url{https://openneuro.org/datasets/ds007020}
\BIBentrySTDinterwordspacing

\bibitem{aljalal2022parkinson}
M.~Aljalal, S.~A. Aldosari, K.~AlSharabi, A.~M. Abdurraqeeb, and F.~A. Alturki, ``Parkinson’s disease detection from resting-state eeg signals using common spatial pattern, entropy, and machine learning techniques,'' \emph{Diagnostics}, vol.~12, no.~5, p. 1033, 2022.

\bibitem{roy2019deep}
Y.~Roy, H.~Banville, I.~Albuquerque, A.~Gramfort, T.~H. Falk, and J.~Faubert, ``Deep learning-based electroencephalography analysis: a systematic review,'' \emph{Journal of Neural Engineering}, vol.~16, no.~5, p. 051001, 2019.

\bibitem{rockhill2020ucsdEEG}
\BIBentryALTinterwordspacing
A.~P. Rockhill, N.~Jackson, J.~George, A.~R. Aron, and N.~C. Swann, ``Uc san diego resting-state eeg data from patients with parkinson’s disease,'' 2020. [Online]. Available: \url{https://openneuro.org/datasets/ds002778}
\BIBentrySTDinterwordspacing

\bibitem{singh2018parkinsons}
\BIBentryALTinterwordspacing
A.~Singh, S.~P. Richardson, N.~Narayanan, and J.~F. Cavanagh, ``Mid-frontal theta activity is diminished during cognitive control in parkinson's disease,'' \emph{Neuropsychologia}, vol. 117, pp. 113--122, 2018. [Online]. Available: \url{https://www.sciencedirect.com/science/article/pii/S0028393218302185}
\BIBentrySTDinterwordspacing

\bibitem{singh2020frontal}
A.~Singh, R.~C. Cole, A.~I. Espinoza, D.~Brown, J.~F. Cavanagh, and N.~S. Narayanan, ``Frontal theta and beta oscillations during lower-limb movement in parkinson’s disease,'' \emph{Clinical Neurophysiology}, vol. 131, no.~3, pp. 694--702, 2020.

\bibitem{brookshire2024data}
G.~Brookshire, J.~Kasper, N.~M. Blauch, Y.~C. Wu, R.~Glatt, D.~A. Merrill, S.~Gerrol, K.~J. Yoder, C.~Quirk, and C.~Lucero, ``Data leakage in deep learning studies of translational eeg,'' \emph{Frontiers in Neuroscience}, vol.~18, p. 1373515, 2024.

\bibitem{roberts2017cross}
D.~R. Roberts, V.~Bahn, S.~Ciuti, M.~S. Boyce, J.~Elith, G.~Guillera-Arroita, S.~Hauenstein, J.~J. Lahoz-Monfort, B.~Schr{\"o}der, W.~Thuiller \emph{et~al.}, ``Cross-validation strategies for data with temporal, spatial, hierarchical, or phylogenetic structure,'' \emph{Ecography}, vol.~40, no.~8, pp. 913--929, 2017.

\bibitem{schratz2019hyperparameter}
P.~Schratz, J.~Muenchow, E.~Iturritxa, J.~Richter, and A.~Brenning, ``Hyperparameter tuning and performance assessment of statistical and machine-learning algorithms using spatial data,'' \emph{Ecological Modelling}, vol. 406, pp. 109--120, 2019.

\bibitem{shafiezadeh2023methodological}
S.~Shafiezadeh, G.~M. Duma, G.~Mento, A.~Danieli, L.~Antoniazzi, F.~Del Popolo~Cristaldi, P.~Bonanni, and A.~Testolin, ``Methodological issues in evaluating machine learning models for eeg seizure prediction: Good cross-validation accuracy does not guarantee generalization to new patients,'' \emph{Applied Sciences}, vol.~13, no.~7, p. 4262, 2023.

\bibitem{ben2010theory}
S.~Ben-David, J.~Blitzer, K.~Crammer, A.~Kulesza, F.~Pereira, and J.~W. Vaughan, ``A theory of learning from different domains,'' \emph{Machine learning}, vol.~79, no.~1, pp. 151--175, 2010.

\bibitem{mansour2008domain}
Y.~Mansour, M.~Mohri, and A.~Rostamizadeh, ``Domain adaptation with multiple sources,'' \emph{Advances in neural information processing systems}, vol.~21, 2008.

\bibitem{torralba2011unbiased}
A.~Torralba and A.~A. Efros, ``Unbiased look at dataset bias,'' in \emph{CVPR 2011}.\hskip 1em plus 0.5em minus 0.4em\relax IEEE, 2011, pp. 1521--1528.

\bibitem{gulrajani2020search}
I.~Gulrajani and D.~Lopez-Paz, ``In search of lost domain generalization,'' \emph{arXiv preprint arXiv:2007.01434}, 2020.

\bibitem{saito2015precision}
T.~Saito and M.~Rehmsmeier, ``The precision-recall plot is more informative than the roc plot when evaluating binary classifiers on imbalanced datasets,'' \emph{PloS one}, vol.~10, no.~3, p. e0118432, 2015.

\bibitem{steyerberg2010assessing}
E.~W. Steyerberg, A.~J. Vickers, N.~R. Cook, T.~Gerds, M.~Gonen, N.~Obuchowski, M.~J. Pencina, and M.~W. Kattan, ``Assessing the performance of prediction models: a framework for traditional and novel measures,'' \emph{Epidemiology}, vol.~21, no.~1, pp. 128--138, 2010.

\bibitem{wang2018deep}
D.~Wang, P.~Cui, and W.~Zhu, ``Deep asymmetric transfer network for unbalanced domain adaptation,'' in \emph{Proceedings of the AAAI Conference on Artificial Intelligence}, vol.~32, no.~1, 2018.

\bibitem{klem1999ten}
G.~H. Klem, H.~O. L{\"u}ders, H.~H. Jasper, and C.~Elger, ``The ten–twenty electrode system of the international federation,'' \emph{Electroencephalography and Clinical Neurophysiology}, vol.~52, pp. 3--6, 1999.

\bibitem{acharya2016acns}
J.~N. Acharya, A.~Hani, P.~Thirumala, and T.~N. Tsuchida, ``American clinical neurophysiology society guideline 3: A proposal for standard montages to be used in clinical eeg,'' \emph{Journal of Clinical Neurophysiology}, vol.~33, no.~4, pp. 312--316, 2016.

\bibitem{schirrmeister2017deep}
R.~T. Schirrmeister, J.~T. Springenberg, L.~D.~J. Fiederer, M.~Glasstetter, K.~Eggensperger, M.~Tangermann, F.~Hutter, W.~Burgard, and T.~Ball, ``Deep learning with convolutional neural networks for eeg decoding and visualization,'' \emph{Human Brain Mapping}, 2017.

\bibitem{samek2017explainable}
W.~Samek, T.~Wiegand, and K.-R. M{\"u}ller, ``Explainable artificial intelligence: Understanding, visualizing and interpreting deep learning models,'' \emph{ITU Journal: ICT Discoveries}, 2017.

\bibitem{anjum2020linear}
M.~F. Anjum, S.~Dasgupta, R.~Mudumbai, A.~Singh, J.~F. Cavanagh, and N.~S. Narayanan, ``Linear predictive coding distinguishes spectral eeg features of parkinson's disease,'' \emph{Parkinsonism \& related disorders}, vol.~79, pp. 79--85, 2020.

\bibitem{shah2020dynamical}
S.~A.~A. Shah, L.~Zhang, and A.~Bais, ``Dynamical system based compact deep hybrid network for classification of parkinson disease related eeg signals,'' \emph{Neural Networks}, vol. 130, pp. 75--84, 2020.

\bibitem{yuvaraj2018novel}
R.~Yuvaraj, U.~Rajendra~Acharya, and Y.~Hagiwara, ``A novel parkinson’s disease diagnosis index using higher-order spectra features in eeg signals,'' \emph{Neural Computing and Applications}, vol.~30, no.~4, pp. 1225--1235, 2018.

\bibitem{lee2019convolutional}
S.~Lee, R.~Hussein, and M.~J. McKeown, ``A deep convolutional-recurrent neural network architecture for parkinson’s disease eeg classification,'' pp. 1--4, 2019.

\bibitem{wu2024multi}
H.~Wu, J.~Qi, E.~Purwanto, X.~Zhu, P.~Yang, and J.~Chen, ``Multi-scale feature and multi-channel selection toward parkinson’s disease diagnosis with eeg,'' \emph{Sensors}, vol.~24, no.~14, p. 4634, 2024.

\bibitem{zhang2025gepd}
Q.~Zhang, R.~Zhang, B.~Zhu, J.~Xiao, Y.~Liu, X.~Han, and Z.~Wang, ``Gepd: Gan-enhanced generalizable model for eeg-based detection of parkinson’s disease,'' in \emph{International Conference on Intelligent Computing}.\hskip 1em plus 0.5em minus 0.4em\relax Springer, 2025, pp. 311--322.

\bibitem{racine2000consistent}
J.~Racine, ``Consistent cross-validatory model-selection for dependent data: hv-block cross-validation,'' \emph{Journal of econometrics}, vol.~99, no.~1, pp. 39--61, 2000.

\bibitem{valavi2018blockcv}
R.~Valavi, J.~Elith, J.~J. Lahoz-Monfort, and G.~Guillera-Arroita, ``blockcv: An r package for generating spatially or environmentally separated folds for k-fold cross-validation of species distribution models,'' \emph{Biorxiv}, p. 357798, 2018.

\bibitem{gemein2020machine}
L.~A.~W. Gemein, R.~T. Schirrmeister, P.~Chrabaszcz, D.~Wilson, J.~Boedecker, A.~Schulze-Bonhage, F.~Hutter, and T.~Ball, ``Machine-learning-based diagnostics of eeg pathology,'' \emph{NeuroImage}, 2020.

\bibitem{rasmussen2025channel}
N.~R. Rasmussen, R.~Rizk, L.~Wang, A.~Singh, and K.~Santosh, ``Channel selected stratified nested cross validation for clinically relevant eeg based parkinsons disease detection,'' \emph{arXiv preprint arXiv:2601.05276}, 2025.

\bibitem{swann2015elevatedsynchrony}
N.~C. Swann, C.~de~Hemptinne, A.~R. Aron, J.~L. Ostrem, R.~T. Knight, and P.~A. Starr, ``Elevated synchrony in parkinson disease detected with electroencephalography,'' \emph{Annals of Neurology}, vol.~78, no.~5, pp. 742--750, Nov. 2015.

\bibitem{stuart2021brain}
S.~Stuart, J.~Wagner, S.~Makeig, and M.~Mancini, ``Brain activity response to visual cues for gait impairment in parkinson’s disease: an eeg study,'' \emph{Neurorehabilitation and neural repair}, vol.~35, no.~11, pp. 996--1009, 2021.

\bibitem{rasmussen2024deepwhalenet}
N.~Rasmussen, R.~Rizk, O.~Matoo, and K.~Santosh, ``Deepwhalenet: climate change-aware fft-based deep neural network for passive acoustic monitoring,'' \emph{International Journal of Pattern Recognition and Artificial Intelligence}, vol.~38, no.~14, p. 2459014, 2024.

\bibitem{rasmussen2025ecologically}
N.~R. Rasmussen, R.~Rizk, L.~Wang, and K.~Santosh, ``Ecologically valid benchmarking and adaptive attention: Scalable marine bioacoustic monitoring,'' \emph{arXiv preprint arXiv:2509.04682}, 2025.

\bibitem{hestness2017deep}
J.~Hestness, S.~Narang, N.~Ardalani, G.~Diamos, H.~Jun, H.~Kianinejad, M.~M.~A. Patwary, Y.~Yang, and Y.~Zhou, ``Deep learning scaling is predictable, empirically,'' \emph{arXiv preprint arXiv:1712.00409}, 2017.

\bibitem{belkin2019reconciling}
M.~Belkin, D.~Hsu, S.~Ma, and S.~Mandal, ``Reconciling modern machine-learning practice and the classical bias--variance trade-off,'' \emph{Proceedings of the National Academy of Sciences}, vol. 116, no.~32, pp. 15\,849--15\,854, 2019.

\bibitem{pan2009survey}
S.~J. Pan and Q.~Yang, ``A survey on transfer learning,'' \emph{IEEE Transactions on knowledge and data engineering}, vol.~22, no.~10, pp. 1345--1359, 2009.

\bibitem{yosinski2014transferable}
J.~Yosinski, J.~Clune, Y.~Bengio, and H.~Lipson, ``How transferable are features in deep neural networks?'' \emph{Advances in neural information processing systems}, vol.~27, 2014.

\bibitem{pernet2020issues}
C.~Pernet, M.~I. Garrido, A.~Gramfort, N.~Maurits, C.~M. Michel, E.~Pang, R.~Salmelin, J.~M. Schoffelen, P.~A. Valdes-Sosa, and A.~Puce, ``Issues and recommendations from the ohbm cobidas meeg committee for reproducible eeg and meg research,'' \emph{Nature neuroscience}, vol.~23, no.~12, pp. 1473--1483, 2020.

\bibitem{blankertz2011single}
B.~Blankertz, S.~Lemm, M.~Treder, S.~Haufe, and K.-R. M{\"u}ller, ``Single-trial analysis and classification of erp components—a tutorial,'' \emph{NeuroImage}, vol.~56, no.~2, pp. 814--825, 2011.

\bibitem{arjovsky2019invariant}
M.~Arjovsky, L.~Bottou, I.~Gulrajani, and D.~Lopez-Paz, ``Invariant risk minimization,'' \emph{arXiv preprint arXiv:1907.02893}, 2019.

\bibitem{guyon2003introduction}
I.~Guyon and A.~Elisseeff, ``An introduction to variable and feature selection,'' \emph{Journal of machine learning research}, vol.~3, no. Mar, pp. 1157--1182, 2003.

\bibitem{S0219635225009532}
S.~Roy, J.~Nuamah, T.~J. Bosch, R.~Barsainya, M.~Scherer, T.~Koeglsperger, K.~Santosh, and A.~Singh, ``Eeg-based classification of parkinson’s disease with freezing of gait using midfrontal beta oscillations,'' \emph{J. Integr. Neurosci.}, vol.~24, no.~6, 2025.

\bibitem{gabeff2021interpreting}
V.~Gabeff, T.~Teijeiro, M.~Zapater, L.~Cammoun, S.~Rheims, P.~Ryvlin, and D.~Atienza, ``Interpreting deep learning models for epileptic seizure detection on eeg signals,'' \emph{Artificial intelligence in medicine}, vol. 117, p. 102084, 2021.

\bibitem{CoPilot}
M.~Corporation, ``Microsoft copilot: Ai companion for productivity and research,'' Online. Available: https://copilot.microsoft.com, 2025, accessed: Oct. 23, 2025. Developed by Microsoft as a large language model-based assistant for writing, coding, and research support.

\end{thebibliography}
}

\clearpage

\clearpage

\begin{table}[t!]
\scriptsize
\centering
\begin{tabular}{lccccccc}
\hline
\textbf{Population (Group)} & \textbf{N} & \textbf{Age} & \textbf{Sex} & \textbf{DD} & \textbf{MS} & \textbf{H\&Y} & \textbf{UPDRS-III} \\
\hline
P1 (PD) & 27 & $69.5 \pm 8.7$ & 10 / 17 & $5.7 \pm 4.2$ & On OR Off & -- & $22.2 \pm 10.3$ \\
P1 (Control) & 27 & $69.5 \pm 9.3$ & 10 / 17 & -- & N/A & -- & -- \\
P2 (PD) & 83 & $68.6 \pm 7.9$ & 58 / 25 & $5.1 \pm 3.9$ & On & $2.1 \pm 0.9$ & $13.4 \pm 7.2$ \\
P2 (Control) & 42 & $71.4 \pm 7.5$ & 24 / 18 & -- & N/A & -- & -- \\
P3 (PD) & 15 & $63.2 \pm 8.2$ & 7 / 8 & $4.5 \pm 3.5$ & On OR Off & 2--3 & -- \\
P3 (Control) & 16 & $63.5 \pm 9.6$ & 7 / 9 & -- & N/A & -- & -- \\
P4 (PD) & 47 & $69.2 \pm 6.8$ & 27 / 18 & 7.8 +/- 5.8 & On & $2.1 \pm 0.3$ & 32.5 +/- 14.4 \\
P4 (Control) & 21 & $69.8 \pm 8.0$ & 11 / 10 & -- & N/A & -- & -- \\
P5 (PD) & 14 & $68.5 \pm 3.7$ & 8 / 6 & $6.6 \pm 3.5$ & On & -- & $29.5 \pm 19.2$ \\
P5 (Control) & 4 & $62.8 \pm 5.6$ & 1 / 3 & -- & N/A & -- & -- \\
\hline
\end{tabular}
\caption{Demographic and clinical characteristics of the included cohorts. Columns report sample size (N), age in years (Age; mean $\pm$ standard deviation), sex distribution (Sex; male/female), disease duration in years (DD; mean $\pm$ standard deviation), medication state at acquisition (MS), Hoehn and Yahr stage (H\&Y), and motor subsection of the Movement Disorder Society Unified PD Rating Scale (UPDRS-III). Missing entries indicate unavailable metadata.}
\label{patientMetadata}
\end{table}

\clearpage

\clearpage

\clearpage

\begin{table}[h]
\scriptsize
\centering
\begin{tabular}{ccccccccc}
\hline
Dataset & Patients & Patients (0/1) & Samples / Patients & Samples & Samples (0/1) & EEG Minutes / Patient \\
\hline
P1 & 56 & 29 / 27 & 4{,}221 & $75.4 \pm 40.6$ & 2268 / 1953 &  $3.23\;[1.22,\,10.07]$ \\

P2 & 122 & 41 / 81 & 7{,}960 & $65.2 \pm 15.8$ & 2690 / 5270 &  $2.48\;[2.01,\,5.71]$ \\

P3 & 31 & 16 / 15 & 1{,}024 & $33.0 \pm 5.7$ & 512 / 512 &  $3.17\;[3.02,\,4.82]$ \\

P4 & 59 & 21 / 38 & 3{,}186 & $54.0 \pm 0.0$ & 1134 / 2052 &  $5.16\;[4.64,\,6.21]$ \\

P5 & 17 & 4 / 13 & 1{,}701 & $100.1 \pm 32.0$ & 378 / 1323 &  $4.32\;[4.12,\,4.96]$ \\
\hline
\end{tabular}
\caption{Dataset quantification after preprocessing and windowing. Reported values include patient counts, patient- and sample-level class balance, total number of generated samples, fraction of total samples contributed by each dataset, mean number of samples per patient (with standard deviation), and median effective EEG recording duration per patient (with minimum and maximum).}
\label{tab:dataset_quantification}
\end{table}

\clearpage

\clearpage

\newcommand{\best}[1]{\bfseries #1}
\begin{table*}[h]
\tiny
\centering
\begin{tabularx}{\textwidth}{|>{\columncolor{gray!15}}c|XXXXXl XXXXXl XXXXXl XXXXXl|}
\toprule
\rowcolor{gray!35}
\multicolumn{1}{|c}{} & \multicolumn{6}{|c}{\textbf{Baseline}} & \multicolumn{6}{|c}{\textbf{Two Channels}} & \multicolumn{6}{|c}{\textbf{Four Channels}} & \multicolumn{6}{|c|}{\textbf{Eight Channels}} \\
\midrule
\rowcolor{gray!15}
\textbf{N-Gram} & \textbf{Acc} & $\sigma$ & \textbf{Rec} & $\sigma$ & \textbf{Pre} & $\sigma$ & \textbf{Acc} & $\sigma$ & \textbf{Rec} & $\sigma$ & \textbf{Pre} & $\sigma$& \textbf{Acc} & $\sigma$ & \textbf{Rec} & $\sigma$ & \textbf{Pre} & $\sigma$& \textbf{Acc} & $\sigma$ & \textbf{Rec} & $\sigma$ & \textbf{Pre} & $\sigma$ \\
\midrule
\rowcolor{gray!15}\multicolumn{1}{|c}{} & \multicolumn{24}{|c|}{\textbf{P1}} \\

1 &
\textbf{.567} & .093 & .917 & .106 & \textbf{.542} & .072 &
.545 & \textbf{.047} & .907 & .088 & .519 & \textbf{.033} &
\textbf{.567} & .053 & \textbf{.944} & \textbf{.048} & .531 & \textbf{.033} & 
\textbf{.567} & .078 & \textbf{.944} & .088 & .535 & .050 \\

2 &
\textbf{.658} & .061 & \textbf{.815} & .119 & \textbf{.617} & .069 & 
.634 & \textbf{.057} & .796 & \textbf{.073} & .595 & \textbf{.055} &
.640 & .065 & .796 & .077 & .605 & .070 & 
.640 & .079 & .790 & .077 & .606 & .080 \\

3 &
.710 & .073 & .750 & \textbf{.035} & \textbf{.698} & .118 & 
\textbf{.723} & .059 & \textbf{.815} & .179 & .691 & \textbf{.081} &
.692 & .067 & .759 & .202 & .671 & .083 & 
.692 & \textbf{.047} & .731 & .126 & .679 & .087 \\

4 &
.571 & N/A & .593 & N/A & .552 & N/A & 
.589 & N/A & .593 & N/A & .571 & N/A &
\textbf{.625} & N/A & \textbf{.667} & N/A & \textbf{.600} & N/A & 
.607 & N/A & .630 & N/A & .586 & N/A \\

\midrule
\rowcolor{gray!15}\multicolumn{1}{|c}{} & \multicolumn{24}{|c|}{\textbf{P2}} \\

1 &
\textbf{.703} & .019 & .898 & .054 & \textbf{.725} & .035 &
.684 & .058 & .883 & .054 & .713 & .047 &
.697 & .033 & .917 & .032 & .712 & .032 & 
.701 & \textbf{.024} & \textbf{.920} & \textbf{.029} & .714 & \textbf{.025} \\

2 &
.734 & .029 & \textbf{.893} & .069 & .756 & .043 & 
.721 & .028 & .872 & .068 & .750 & .021 &
\textbf{.739} & \textbf{.020} & .879 & \textbf{.058} & \textbf{.765} & \textbf{.017} & 
.727 & .030 & .864 & .073 & .760 & .023 \\

3 &
\textbf{.711} & \textbf{.026} & \textbf{.778} & \textbf{.050} & \textbf{.787} & \textbf{.039} & 
.691 & .055 & .744 & .130 & .785 & .043 &
.684 & .039 & .731 & .114 & .785 & .040 & 
.693 & .034 & .750 & .099 & .784 & .043 \\

4 &
\textbf{.770} & N/A & .951 & N/A & \textbf{.762} & N/A & 
.672 & N/A & \textbf{1.000} & N/A & .669 & N/A &
.705 & N/A & .988 & N/A & .696 & N/A & 
.713 & N/A & .988 & N/A & .702 & N/A \\

\midrule
\rowcolor{gray!15}\multicolumn{1}{|c}{} & \multicolumn{24}{|c|}{\textbf{P3}} \\

1 &
.621 & .081 & .650 & .257 & .660 & \textbf{.140} &
\textbf{.677} & \textbf{.070} & \textbf{.683} & \textbf{.240} & \textbf{.746} & .201 &
.661 & .081 & .667 & .261 & .728 & .177 & 
.653 & .096 & .667 & .261 & .723 & .183 \\

2 &
.629 & .034 & .544 & .129 & \textbf{.658} & .084 & 
\textbf{.645} & \textbf{.041} & \textbf{.611} & .176 & .655 & .063 &
.634 & .049 & .600 & .193 & .637 & \textbf{.058} & 
.629 & .057 & .578 & \textbf{.124} & .648 & .094 \\

3 &
.629 & \textbf{.019} & .450 & .084 & .696 & .098 & 
\textbf{.710} & .026 & \textbf{.633} & .086 & \textbf{.748} & .100 &
.694 & .032 & .600 & \textbf{.077} & .724 & \textbf{.056} & 
.694 & .042 & .583 & .137 & .744 & .092 \\

4 &
.645 & N/A & .333 & N/A & .833 & N/A & 
\textbf{.742} & N/A & \textbf{.533} & N/A & \textbf{.889} & N/A &
.710 & N/A & .467 & N/A & .875 & N/A & 
.710 & N/A & .467 & N/A & .875 & N/A \\

\midrule
\rowcolor{gray!15}\multicolumn{1}{|c}{} & \multicolumn{24}{|c|}{\textbf{P4}} \\

1 &
\textbf{.674} & .040 & .868 & .134 & \textbf{.718} & .102 & 
.648 & .042 & \textbf{.882} & .143 & .685 & \textbf{.073} &
.661 & \textbf{.034} & \textbf{.882} & \textbf{.126} & .698 & .081 & 
.661 & .046 & .875 & .135 & .706 & .107 \\

2 &
.667 & .033 & .746 & .196 & .773 & .103 & 
.672 & \textbf{.026} & .746 & .184 & .772 & \textbf{.086} &
\textbf{.689} & .044 & \textbf{.776} & \textbf{.173} & \textbf{.783} & .108 & 
.678 & .047 & .763 & .178 & .781 & .115 \\

3 &
.691 & .035 & .750 & .117 & .779 & .075 & 
.669 & .045 & .724 & .137 & .767 & .080 &
.682 & \textbf{.025} & .711 & \textbf{.107} & .786 & \textbf{.061} & 
\textbf{.708} & .029 & \textbf{.757} & .110 & \textbf{.794} & .072 \\

4 &
.729 & N/A & .684 & N/A & \textbf{.867} & N/A & 
\textbf{.763} & N/A & \textbf{.763} & N/A & .853 & N/A & 
.746 & N/A & \textbf{.763} & N/A & .829 & N/A & 
.729 & N/A & .737 & N/A & .824 & N/A \\

\midrule
\rowcolor{gray!15}\multicolumn{1}{|c}{} & \multicolumn{24}{|c|}{\textbf{P5}} \\

1 &
.824 & .068 & .962 & .044 & .839 & .070 & 
.809 & \textbf{.056} & .962 & .044 & .822 & \textbf{.048} &
\textbf{.838} & .088 & \textbf{.981} & \textbf{.038} & \textbf{.840} & .073 & 
.824 & .068 & .962 & .044 & .839 & .070 \\

2 &
\textbf{.902} & .080 & .949 & .079 & \textbf{.928} & .060 & 
.863 & \textbf{.030} & .910 & .058 & .914 & \textbf{.051} &
\textbf{.902} & .061 & \textbf{.987} & \textbf{.031} & .900 & .065 & 
.892 & .078 & .974 & .040 & .897 & .069 \\

3 &
.897 & .100 & .923 & .109 & .940 & .041 & 
.897 & \textbf{.029} & .942 & \textbf{.038} & .928 & .055 &
.912 & .076 & .942 & .074 & .942 & \textbf{.039} & 
\textbf{.941} & .048 & \textbf{.962} & .044 & \textbf{.963} & .043 \\

4 &
\textbf{.824} & N/A & \textbf{.769} & N/A & \textbf{1.00} & N/A & 
.706 & N/A & .692 & N/A & .900 & N/A &
.706 & N/A & .692 & N/A & .900 & N/A & 
\textbf{.824} & N/A & \textbf{.769} & N/A & \textbf{1.00} & N/A \\

\bottomrule
\end{tabularx}
\caption{Performance comparison across channel selection strategies, and n‑gram levels for each dataset. Results report mean accuracy (Acc), recall (Rec), and precision (Pre) with standard deviation \(\sigma\). Bold values indicate the best metric within each row.}
\label{tab:ab}
\end{table*}

\clearpage

\clearpage

\begin{table}[t!]
\scriptsize
\centering
\begin{tabularx}{\linewidth}{|>{\columncolor{gray!15}}l|XXXXl|XXXXX|}
\toprule
\rowcolor{gray!35}
\multicolumn{1}{|c}{} & \multicolumn{5}{|c}{\textbf{Baseline}} & \multicolumn{5}{|c|}{\textbf{Two Channels}} \\
\midrule
\rowcolor{gray!15}
\textbf{Set} & \textbf{P1} & \textbf{P2} & \textbf{P3} & \textbf{P4} & \textbf{P5} & \textbf{P1} & \textbf{P2} & \textbf{P3} & \textbf{P4} & \textbf{P5} \\
\midrule
\textbf{P1} & N/A & .727 & .632 & .709 & .877 & N/A & .696 & .696 & .698 & .838 \\
\textbf{P2} & .658 & N/A & .656 & .694 & .882 & .652 & N/A & .707 & .679 & .824 \\
\textbf{P3} & .606 & .729 & N/A & .675 & .838 & .601 & .692 & N/A & .688 & .809 \\
\textbf{P4} & .637 & .727 & .599 & N/A & .848 & .626 & .671 & .667 & N/A & .804 \\
\textbf{P5} & .606 & .735 & .637 & .682 & N/A & .612 & .710 & .704 & .688 & N/A \\

\bottomrule
\rowcolor{gray!35}
\multicolumn{1}{|c}{} & \multicolumn{5}{|c}{\textbf{Four Channels}} & \multicolumn{5}{|c|}{\textbf{Eight Channels}} \\
\midrule
\rowcolor{gray!15}
\textbf{Set} & \textbf{P1} & \textbf{P2} & \textbf{P3} & \textbf{P4} & \textbf{P5} & \textbf{P1} & \textbf{P2} & \textbf{P3} & \textbf{P4} & \textbf{P5} \\
\midrule
\textbf{P1} & N/A & .706 & .675 & .710 & .853 & N/A & .708 & .685 & .718 & .887 \\
\textbf{P2} & .650 & N/A & .696 & .688 & .873 & .656 & N/A & .696 & .686 & .897 \\
\textbf{P3} & .607 & .706 & N/A & .689 & .809 & .603 & .705 & N/A & .688 & .843 \\
\textbf{P4} & .650 & .695 & .648 & N/A & .824 & .644 & .705 & .629 & N/A & .853 \\
\textbf{P5} & .616 & .719 & .680 & .691 & N/A & .603 & .715 & .675 & .684 & N/A \\

\bottomrule
\end{tabularx}
\caption{Directional cross‑population classification accuracies for all train–test dataset pairings under channel selection strategies. Each entry reports accuracy obtained when training on the row dataset and evaluating on the column dataset, preserving train–test directionality.}
\label{tab:asym}
\end{table}

\clearpage

\end{document}